\pdfoutput=1

\documentclass[11pt]{article}

\usepackage{EMNLP2023}

\usepackage{times}
\usepackage{latexsym}
\usepackage{kotex}

\usepackage[T1]{fontenc}

\usepackage[utf8]{inputenc}

\usepackage{microtype}

\usepackage{inconsolata}

\usepackage{graphicx}
\usepackage{booktabs}
\usepackage{multirow}

\usepackage{adjustbox}
\usepackage{arydshln}
\usepackage{amsmath}
\usepackage[utf8]{inputenc}

\usepackage{amssymb}
\usepackage{svg}

\usepackage{longtable}
\usepackage{makecell}
\usepackage{array}
\usepackage{caption}
\usepackage{subcaption}
\usepackage{wrapfig}
\usepackage{float}
\usepackage{scalerel}
\usepackage{colortbl}
\usepackage{tablefootnote}
\usepackage{adjustbox}
\usepackage{dblfloatfix}    
\usepackage[algo2e, linesnumbered, ruled, vlined]{algorithm2e}
\usepackage{algorithm}
\usepackage{algorithmic}
\usepackage{enumitem}
\usepackage{xcolor}
\usepackage{pifont}

\usepackage{tcolorbox}
\tcbuselibrary{skins,breakable}
\definecolor{hylarblue}{RGB}{210,228,248}
\definecolor{sftgreen}{RGB}{220,240,220}
\definecolor{simcotorange}{RGB}{255,237,210}
\definecolor{errorred}{RGB}{240,210,210}
\definecolor{correctgreen}{RGB}{200,235,200}
\newtcolorbox{cbox}[2][]{colback=#2,colframe=gray!35,left=3pt,right=3pt,top=1.5pt,bottom=1.5pt,boxrule=0.4pt,arc=2pt,#1}

\SetKwComment{hltcpline}{}{}

\newcommand{\hltcp}[1]{%
  {\setlength{\fboxsep}{1.5pt}%
   \hltcpline{%
     \colorbox{red!3}{%
       \strut\textcolor{red!50!black}{// #1}%
     }%
   }}%
}

\newcommand{\vcorrect}{\textcolor{green!55!black}{\textbf{[\checkmark]}}}
\newcommand{\vwrong}{\textcolor{red!65!black}{\textbf{[$\times$]}}}

\newcommand{\LblockN}[1]{\colorbox{gray!18}{\scriptsize\texttt{[L$\times$#1]}}}
\usepackage{fvextra}

\newtcolorbox{promptbox}[2][]{
  colback=#2,
  colframe=black!35,
  boxrule=0.3pt,
  arc=2pt,
  left=3pt,
  right=3pt,
  top=2pt,
  bottom=2pt,
  #1
}

\DefineVerbatimEnvironment{PromptVerb}{Verbatim}{
  fontsize=\scriptsize,
  breaklines=true,
  breakanywhere=true,
  baselinestretch=0.88
}

\definecolor{darkblue}{rgb}{0.0,0.0,0.65}
\definecolor{darkred}{rgb}{0.65,0.0,0.0}
\definecolor{darkgreen}{rgb}{0.0,0.5,0.0}
\definecolor{pinegreen}{RGB}{1, 121, 111}
\definecolor{tab:blue}{RGB}{31,119,180}
\definecolor{tab:red}{RGB}{214,39,40}
\definecolor{tab:green}{RGB}{44,160,44}
\definecolor{tab:orange}{RGB}{255,127,14}
\hypersetup{
    colorlinks=true,
    citecolor=darkblue,
    linkcolor=darkblue,
    filecolor=darkblue,
    urlcolor=darkblue,
}

\newcommand{\appref}[1]{\hyperref[#1]{Appendix~\ref*{#1}}}
\newcommand{\secref}[1]{\hyperref[#1]{Section~\ref*{#1}}}
\newcommand{\algoref}[1]{\hyperref[#1]{Algorithm~\ref*{#1}}}

\newcommand{\cmark}{\textcolor{darkgreen}{\ding{51}}}
\newcommand{\xmark}{\textcolor{darkred}{\ding{55}}}

\newcommand{\algo}{\textsc{TAC-CoT}\xspace}

\title{\algo: Tool-Empowered Implicit Chain-of-Thought \\via Interleaving Explicit Tokens}

\author{
Juseon Do$^{1}$ \quad Sangmin Bae$^{2}$ \quad Jeesu Jung$^{1,*}$ \quad Hwanjun Song$^{1,3,*}$ \\
$^{1}$Data Intelligence System Lab, KAIST \\
$^{2}$Optimization and Statistical Inference Lab, KAIST AI \\
$^{3}$Cluvion\\
\texttt{dojuseon@gmail.com} \quad
\texttt{\{bsmn0223, jeesu\_jung, songhwanjun\}@kaist.ac.kr}
}

\begin{document}
\maketitle
\renewcommand{\thefootnote}{\fnsymbol{footnote}}
\footnotetext[1]{Corresponding authors.}

\begin{abstract}
LLM-based agentic systems are increasingly combined with CoT reasoning to solve tasks that require accurate reasoning, external knowledge, and multi-step deliberation. Yet, explicit CoT incurs substantial generation overhead, while implicit CoT, which replaces textual rationales with compact continuous tokens, does not naturally support the explicit communication interface required for agentic tool-use. We propose \algo{}, a tool-empowered adaptive implicit CoT framework that replaces explicit reasoning traces with compact latent reasoning segments while retaining textual tokens only for tool calls, observations, and final answers. This interleaved design enables agentic tool-use without verbalizing every intermediate reasoning step. Experiments on agentic math and QA benchmarks show that \algo{} improves accuracy by up to 8.8 points while reducing model forward passes by up to \(6.8\times\). Furthermore, analyses show that \algo{} improves tool-use quality in both single- and multi-tool settings.
\looseness=-1
\end{abstract}
\begin{figure}[!t]
    \centerline{\includegraphics[width=\columnwidth]{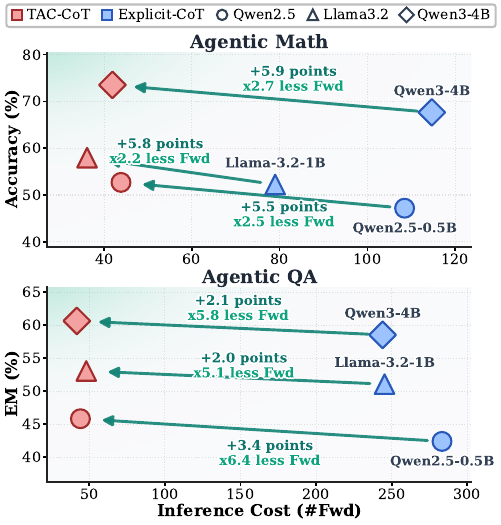}}
    \vspace*{-0.15cm}
    \caption{\algo improves over the Explicit-CoT baseline in both agentic math and question-answering (QA) settings. The two methods use the same tool-using agents and differ only in their reasoning backbone: \algo uses implicit CoT, whereas the baseline uses explicit CoT. \#\,Fwd denotes the number of model forward passes~\cite{kuzina2026kava,wei2026simcot}. 
    \looseness=-1
    }
    \label{fig:pareto}
    \vspace*{-0.4cm}
\end{figure}

\section{Introduction}

To enhance the problem-solving ability of large language models (LLMs), a wide range of reasoning-oriented methods have been explored. 
In particular, for tasks that require accurate and reliable reasoning, such as mathematical reasoning and knowledge-intensive question answering, research has evolved toward Chain-of-Thought (CoT) reasoning, which improves models' internal reasoning capabilities through intermediate reasoning steps~\cite{wei2022chain,deepseekai2025deepseekv32pushingfrontieropen}, as well as agentic tool-use frameworks, which enhance rigor and factuality by enabling models to interact with external tools during problem solving~\cite{zhang2026agentic,kim2026costdynamicreasoningdemystifying}.

\begin{figure*}[!t]
    \centerline{\includegraphics[width=0.85\textwidth]{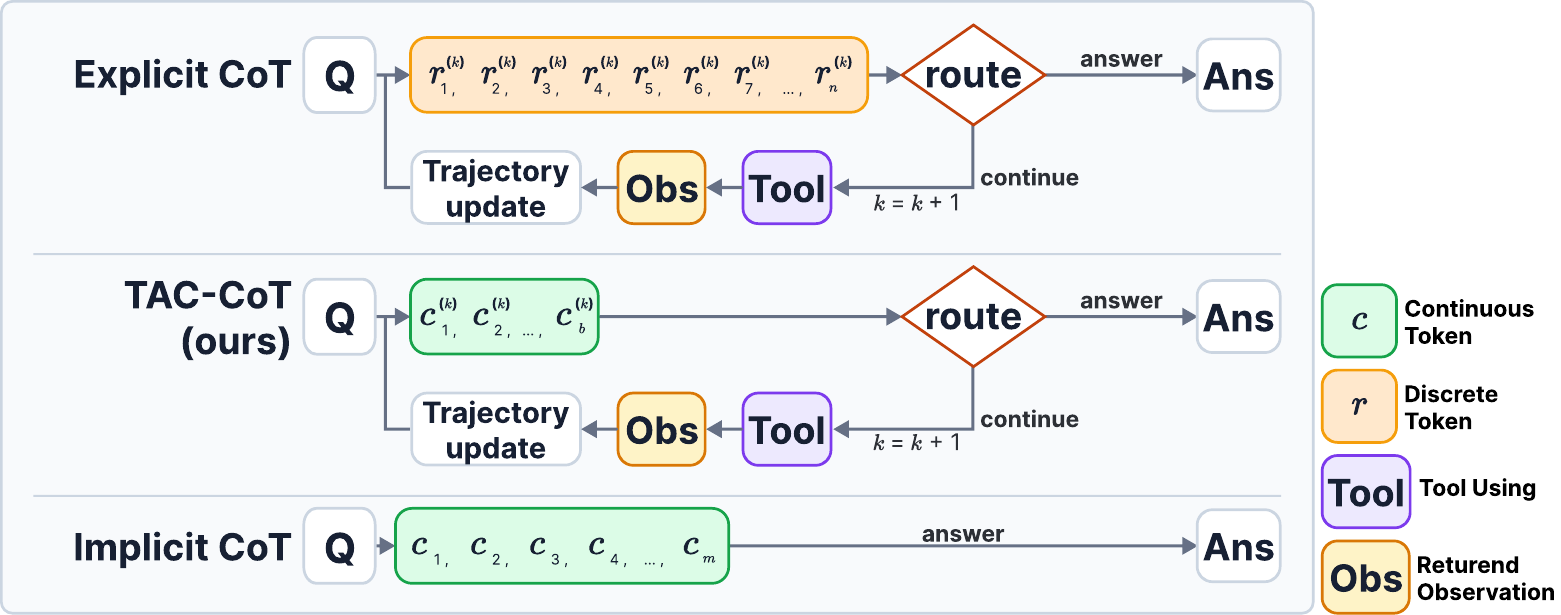}}
    \vspace*{-0.2cm}
    \caption{Overview of \algo compared with the Explicit-CoT agent and implicit-CoT methods. \algo uses implicit CoT as its reasoning backbone while preserving interaction segments in explicit natural language. This design enables implicit CoT to use tools and allocate query-dependent implicit reasoning budgets, improving efficiency, performance, and robustness over the Explicit-CoT agent. Here, $n$ denotes the number of discrete tokens generated by the Explicit-CoT agent at each step, $b$ denotes the number of continuous tokens in each implicit reasoning segment of \algo, and $m$ denotes the fixed continuous-token budget used by prior implicit-CoT methods.
    \looseness=-1
    }
    \vspace*{-0.3cm}
    \label{fig:compare}
\end{figure*}


Most prior work adopts the explicit CoT paradigm, where intermediate reasoning is represented as human-readable natural-language tokens. This paradigm is attractive because it is directly compatible with autoregressive generation and provides an interpretable interface between reasoning and action~\cite{kojima2022large,yao2023tree}. However, verbalizing the entire reasoning process substantially increases inference cost as reasoning traces grow longer and makes the model more susceptible to error accumulation across intermediate steps~\cite{han-etal-2025-token,wu2026when}.

\textit{Implicit} CoT has emerged as an alternative paradigm that replaces readable natural-language rationales with compact \textit{continuous thought tokens}. By performing reasoning in non-verbal latent spaces, implicit CoT aims to retain the benefits of multi-step reasoning while mitigating the inference cost and vulnerability to error accumulation of lengthy textual traces~\cite{shen-etal-2025-codi}.  

Despite these benefits, implicit CoT methods are ill-suited for LLM-based agent systems. They rely on a fixed number of implicit tokens to carry the entire reasoning process of every query, which gives rise to two limitations. First, since the implicit-token budget is consumed in a single uninterrupted block before the final answer \cite{deng2024implicit, wei2026simcot}, there is no intermediate point at which the model can emit an explicit tool call or ingest an observation, so tool use is structurally infeasible. Second, the same token count is applied to every input regardless of its difficulty or the number of intermediate decision points it requires \cite{wu2026when,aggarwal2026optimalthinkingbench}. As a result, prior implicit CoT methods cannot interact with external environments and under-allocate reasoning capacity to more difficult queries. 

In this paper, we propose \algo{} (\underline{T}ool-empowered \underline{A}daptive \underline{C}ontinuous CoT), which \emph{interleaves compact implicit reasoning segments with explicit tokens for tool calls, observations, and final answers}. Rather than carrying the entire reasoning process in a single block of implicit tokens, \algo{} reasons within a short implicit segment and then either invokes a tool or produces the answer, resuming with another segment after each tool observation. This interleaving simultaneously addresses both limitations of prior implicit CoT, as explicit tokens between segments make tool use executable, while the number of segments, determined on the fly, makes the reasoning budget query-adaptive. To train this behavior, we automatically construct tool-use trajectories from reasoning datasets and supervise the model to perform compact latent reasoning between tool invocations.

Our contributions are summarized as follows.\vspace{3pt}\\
\textbf{(1)} We {extend implicit CoT to tool-empowered agentic systems}, enabling compact continuous reasoning in iterative reasoning--action workflows. \vspace{3pt}\\
\textbf{(2)} We show that {\algo improves reasoning performance} while substantially reducing the number of reasoning tokens compared with explicit CoT-style reasoning by interleaving continuous thought tokens with explicit tokens.\vspace{3pt}\\
\textbf{(3)} We introduce {an adaptive token-allocation mechanism} that flexibly adjusts the length of implicit reasoning according to query difficulty.\vspace{3pt}\\
\textbf{(4)} As highlighted in Figure~\ref{fig:pareto}, \algo improves downstream task performance while using the forward reasoning budget more efficiently. Our analysis shows that \algo benefits from query-adaptive implicit reasoning budgets, improves tool-using scenarios, and mitigates error accumulation through implicit reasoning.


\looseness=-1

\section{\algo: Tool-Empowered Implicit CoT via Interleaving Explicit Tokens}
\label{sec:icoffea}


We explain \emph{\algo}, a tool-empowered implicit CoT framework that interleaves latent reasoning segments with explicit, executable tool interactions. We first describe \algo{}'s inference procedure (\S\ref{subsec:method_21}), then how we construct training trajectories (\S\ref{sec:data}), and finally the learning objective (\S\ref{sec:training}).
Algorithm~\ref{alg:tac-cot-inference} shows its inference procedure.

\subsection{Tool-Interleaved Implicit CoT with Query-Adaptive Budgets}
\label{subsec:method_21}

The key idea of \algo is to jointly leverage implicit reasoning tokens and textual tokens for tool-use trajectories. In this process, the length of the reasoning tokens is adaptively adjusted according to the query. As shown in Figure~\ref{fig:compare}, this produces an interleaved trajectory that preserves the executable interface of tool-using agents without verbalizing every intermediate step.

\looseness=-1

Given a query $q$, we consider an \emph{explicit} tool-using CoT trajectory as:
\begin{equation}
\tau_E = \bigl(\mathcal{R}^{(1)}, u^{(1)}, o^{(1)}, \ldots, \mathcal{R}^{(k)}, a\bigr),
\end{equation}
where $\mathcal{R}^{(i)} = (r^{(i)}_1, \ldots, r^{(i)}_{n})$ is the $i$-th explicit reasoning segment of $n$ discrete tokens; $u^{(i)}$ and $o^{(i)}$ are the corresponding tool call and observation, and $a$ is the final answer. The last segment $\mathcal{R}^{(k)}$ produces no tool call, as the model uses it to consolidate prior observations before generating $a$.
\algo{} constructs the corresponding \emph{interleaving} trajectory $\tau_{I}$ by replacing each explicit reasoning segment $\mathcal{R}^{(i)}$ with an implicit segment $C^{(i)} = (c^{(i)}_1, \ldots, c^{(i)}_b)$ of $b$~($<n$) continuous thought tokens, while keeping $u^{(i)}$, $o^{(i)}$, and $a$ in their original explicit textual form:
\begin{equation}
\tau_I = \bigl(C^{(1)}, u^{(1)}, o^{(1)}, \ldots, C^{(k)}, a\bigr).
\label{eq:interleaving}
\end{equation}
Each continuous thought token $c^{(i)}_t$ is obtained by applying a projection layer to the last-layer hidden state at the current position and is fed back to the model via input embeddings, bypassing the vocabulary projection; following CODI \cite{shen-etal-2025-codi}, this projection layer separates the latent thought space from the token embedding space.



\begin{algorithm2e}[t]
\caption{TAC-CoT inference}
\label{alg:tac-cot-inference}
\DontPrintSemicolon
\SetKw{KwRet}{return}

\KwIn{Query $q$, model $f_\theta$, tool executor $\mathcal{E}$,
implicit segment length $b$}
\KwOut{Final answer $a$}

$\tau \leftarrow [\,]$\tcp*[r]{prefix of $\tau_I$}

\Repeat{$\ell_{\mathrm{ans}} \ge \ell_{\mathrm{act}}$}{
    \hltcp{Implicit reasoning segment}
    \For{$t \leftarrow 1$ \KwTo $b$}{
        $h_t \leftarrow \mathrm{LastHidden}_\theta(q \circ \tau)$\;
        \hltcp{Continuous thought token}
        $c_t \leftarrow \mathrm{Proj}_\theta(h_t)$\;
        $\tau \leftarrow \tau \circ [c_t]$\;
    }

    \hltcp{Route selection}
    $(\ell_{\mathrm{act}}, \ell_{\mathrm{ans}})
    \leftarrow \mathrm{RouteLogits}_\theta(q \circ \tau)$\;

    \If{$\ell_{\mathrm{act}} > \ell_{\mathrm{ans}}$}{
        $u \leftarrow \mathrm{DecodeTool}_\theta(q \circ \tau)$\;
        $o \leftarrow \mathcal{E}(u)$\;
        $\tau \leftarrow \tau \circ [u,o]$\;
    }
}

$a \leftarrow \mathrm{DecodeAns}_\theta(q \circ \tau)$\;
\KwRet{$a$}\;
\end{algorithm2e}

\paragraph{Deciding When to Call a Tool.} 
A key design decision of \algo{} is \emph{when} to interrupt latent reasoning for an external interaction. At the end of every implicit segment $C^{(i)}$, \algo{} compares the logits of the tool call $u^{(i)}$ and answer options to decide the next step. If the tool-call logit is higher, \algo{} emits a tool call, appends the returned observation $o^{(i)}$ to the context, and starts another implicit segment. Otherwise, the current segment becomes the consolidating segment $C^{(k)}$, from which the final answer is generated. Unlike prior implicit CoT methods that consume a fixed implicit budget before any explicit output, this per-segment routing is what makes tool use feasible under implicit reasoning.

\paragraph{Query-Adaptive Continuous-Thought Token Budget.}
Since the tool call decision is made after every implicit segment, the total reasoning budget becomes \emph{query-adaptive}. Simple queries may terminate after a single segment, while more complex queries can continue through multiple tool interactions. Thus, $b$ is fixed only locally within each segment. In contrast, if \algo{} terminates after $k$ implicit segments $C^{(i)}$, it uses $kb$ continuous tokens in total, where $k$ is determined dynamically during inference.

\looseness=-1

\subsection{Dataset Construction for Agentic CoT}
\label{sec:data}



Realizing our interleaved inference mechanism in Eq~\eqref{eq:interleaving} requires training data that exposes the explicit counterpart $\tau_E$ of the interleaved trajectory $\tau_I$. That is, each example must provide the full sequence of explicit reasoning segments $\mathcal{R}^{(i)}$, tool calls $u^{(i)}$, observations $o^{(i)}$, and the final answer $a$, from which \algo{} learns to replace each $\mathcal{R}^{(i)}$ with an implicit segment $C^{(i)}$ while preserving the surrounding explicit interactions.

We construct such data with two strategies depending on whether the source dataset already includes executable tool interactions, or only reasoning traces and gold answers.
\vspace*{-0.15cm}
\begin{itemize}[leftmargin=*]
    \item \underline{\emph{Sources with Tool Interactions.}} We convert the provided executable annotations into tool calls. If the executable annotations are provided separately from the reasoning traces, we match their tool-call parameters and returned observations to the corresponding reasoning spans. We then normalize the matched interactions into a unified tool-call format with explicit tool names and parameters, and insert the tool calls and observations into the reasoning trajectory.
    \vspace*{-0.15cm}
    \item \underline{\emph{Sources without Tool Interactions.}} As intermediate interactions are not directly available, we instead employ a stronger LLM (DeepSeek-V3; See Appendix \ref{sec:appendix-data}) as a tool-augmented solver to synthesize them. Given a question and the available tools, the model generates a complete trajectory with interleaved reasoning spans, tool calls, observations, and a predicted answer. We retain only trajectories whose predicted answers match the source gold answers, and normalize the retained trajectories into the same format.
\end{itemize}

\looseness=-1


\begin{figure}[!t]
    \vspace*{-0.3cm}
    \centerline{\includegraphics[width=\columnwidth]{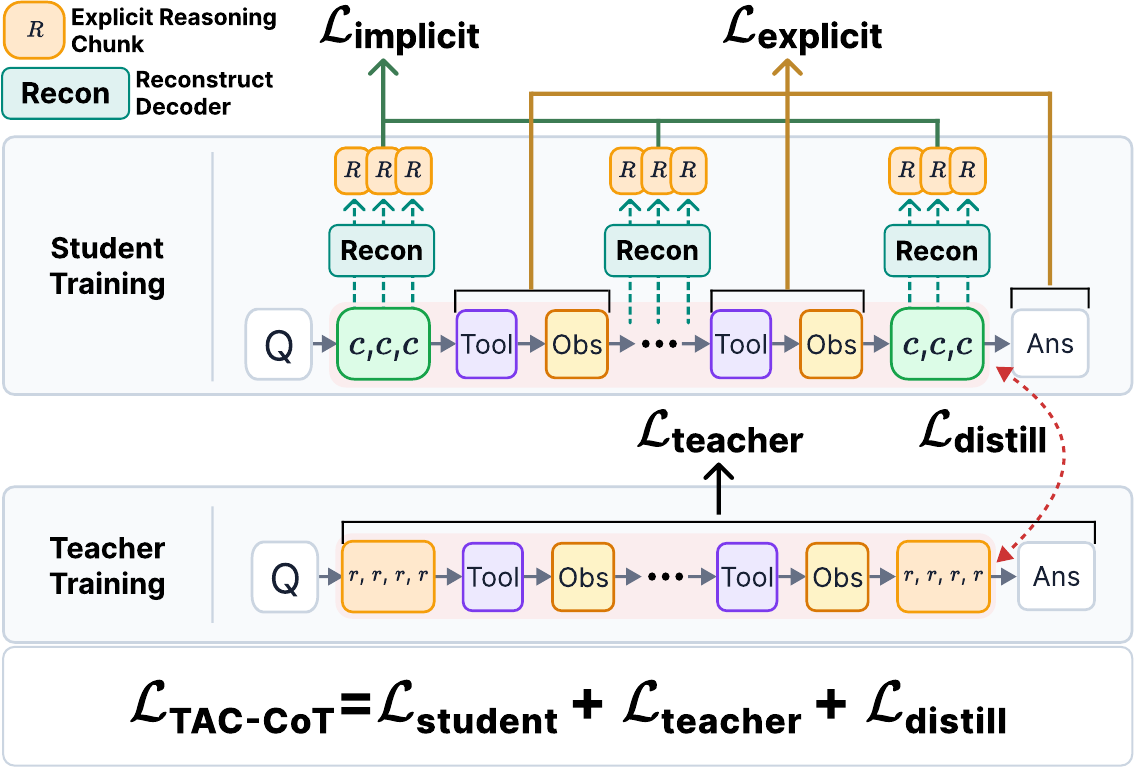}}
    \caption{Training objective of \algo. The student generates continuous thought tokens, from which explicit reasoning chunks are reconstructed for supervision. Explicit outputs are trained with cross-entropy loss, while answer-boundary hidden states are distilled from an explicit-CoT teacher. The final objective combines student, teacher, and distillation losses.
    }
    \label{fig:training}
    \vspace*{-0.3cm}
\end{figure}


\subsection{Training Objective}
\label{sec:training}

Given the explicit trajectory $\tau_E$ from \secref{sec:data}, training must teach the model to (i) encode the reasoning of each $\mathcal{R}^{(i)}$ into an implicit segment $C^{(i)}$, and (ii) preserve the surrounding tool calls $u^{(i)}$, observations $o^{(i)}$, and final answer $a$ in explicit form. We achieve both through self-distillation~\cite{shen-etal-2025-codi}, where the same backbone model serves as both the student and the teacher, with $\tau_E$ providing the teacher signal, as in Figure \ref{fig:training}.


\paragraph{Student Training.}
Following the reconstruction idea of \citet{wei2026simcot}, we attach an auxiliary reconstruction decoder \(D_\phi\), which is discarded at inference.
Because \algo hides only reasoning spans, the auxiliary decoder reconstructs the omitted reasoning content rather than the whole trajectory.
Each \(\mathcal{R}^{(i)}\) is split into \(b\) chunks \((R^{(i)}_1,\ldots,R^{(i)}_b)\), and each continuous thought token \(c^{(i)}_t\) is aligned with \(R^{(i)}_t\):
\begin{equation}
\mathcal{L}_{\mathrm{implicit}}
=
-\sum_{i=1}^{K}\sum_{t=1}^{B}
\log q_\phi\!\left(R^{(i)}_t \mid c^{(i)}_t\right),
\end{equation}
where \(q_\phi(R^{(i)}_t \mid c^{(i)}_t)\) is the autoregressive reconstruction likelihood assigned by \(D_\phi\) to \(R^{(i)}_t\).
This loss forces each implicit segment to encode the reasoning content that would otherwise be verbalized at the corresponding point in the tool-use trajectory.

For the model-generated explicit segments, including tool calls and final answers, we apply the standard cross-entropy loss \(\mathcal{L}_{\mathrm{explicit}}\).
Observations are kept as explicit context returned by the environment.
The student loss is
\begin{equation}
\mathcal{L}_{\mathrm{student}}
=
\mathcal{L}_{\mathrm{implicit}}
+
\mathcal{L}_{\mathrm{explicit}}.
\end{equation}

\paragraph{Teacher Training and Distillation.}
We train the teacher with the standard cross-entropy loss \(\mathcal{L}_{\mathrm{teacher}}\) on the explicit trajectory. 
We then distill the teacher's hidden activations at the answer boundary so that the student after implicit reasoning matches the teacher after explicit reasoning:
\begin{equation}
\!\!\mathcal{L}_{\mathrm{distill}}
=
\frac{1}{M}
\sum_{l=1}^{M}
\left\|
\operatorname{sg}\!\left[
h^{l}_{\mathrm{teacher}}
\right]
-
h^{l}_{\mathrm{student}}
\right\|_{1},\!\!
\end{equation}
\(h^{l}_{\mathrm{teacher}}\) and \(h^{l}_{\mathrm{student}}\) are the teacher and student hidden activations at the colon token before answer generation, \(M\) is the number of layers, and \(\operatorname{sg}[\cdot]\) denotes the stop-gradient operation.
The final objective is therefore \(\mathcal{L}_{\algo}=\mathcal{L}_{\mathrm{student}}+\mathcal{L}_{\mathrm{teacher}}+\mathcal{L}_{\mathrm{distill}}\).

\begin{table}[t]
\centering
\small
\setlength{\tabcolsep}{2pt}
\resizebox{\linewidth}{!}{
\begin{tabular}{l|ll}
\toprule
\textbf{Task} & \textbf{Benchmark} & \textbf{Setting} \\
\midrule
\multirow{3}{*}{Math}
& GSM8K~\cite{cobbe2021trainingverifierssolvemath} & In-domain \\
& GSM-Hard~\cite{10.5555/3618408.3618843} & Out-of-domain \\
& SVAMP~\cite{patel-etal-2021-nlp} & Out-of-domain \\
\midrule
\multirow{3}{*}{QA}
& HotpotQA~\cite{yang-etal-2018-hotpotqa} & In-domain \\
& 2WikiMultihopQA~\cite{ho-etal-2020-constructing} & Out-of-domain \\
& KILT-TriviaQA~\cite{petroni-etal-2021-kilt} & Out-of-domain \\
\midrule
Tool-use
& ToolBench~\cite{qin2024toolllm} & Tool-use evaluation \\
\bottomrule
\end{tabular}
}
\caption{Evaluation benchmarks for agentic mathematical reasoning, agentic QA and tool-use capability.}
\label{tab:evaluation_benchmarks}
\end{table}

\begin{table*}[t]
\centering
\setlength{\tabcolsep}{3pt}
\resizebox{\textwidth}{!}{

\begin{tabular}{lc |c cc c cccc c cc c|c cc c ccc c ccc}
\toprule
&&&\multicolumn{10}{c}{\textbf{Math}}&&\multicolumn{11}{c}{\textbf{Question-Answering}}\\
\cmidrule(l{2pt}r{2pt}){3-13} \cmidrule(l{2pt}r{2pt}){13-25}
\multirow{2.5}{*}{\textbf{Method}} & \multirow{2.5}{*}{\textbf{TU}} && \multicolumn{2}{c}{\textbf{In-domain}} & & \multicolumn{4}{c}{\textbf{Out-of-domain}} &  & &&&& \multicolumn{2}{c}{\textbf{In-domain}} && \multicolumn{4}{c}{\textbf{Out-of-domain}} & & &\\
\cmidrule(l{2pt}r{2pt}){4-5} \cmidrule(l{2pt}r{2pt}){7-10} \cmidrule(l{2pt}r{2pt}){16-17} \cmidrule(l{2pt}r{2pt}){19-22}
&&& \multicolumn{2}{c}{GSM8K} && \multicolumn{2}{c}{GSM-Hard} & \multicolumn{2}{c}{SVAMP} &  &\multicolumn{2}{c}{Average}&&& \multicolumn{2}{c}{HotpotQA} && \multicolumn{2}{c}{2WikiMulti} & \multicolumn{2}{c}{TriviaQA} & & \multicolumn{2}{c}{Average}\\
\cmidrule(l{2pt}r{2pt}){4-5} \cmidrule(l{2pt}r{2pt}){7-8} \cmidrule(l{2pt}r{2pt}){9-10} \cmidrule(l{2pt}r{2pt}){12-13} \cmidrule(l{2pt}r{2pt}){16-17} \cmidrule(l{2pt}r{2pt}){19-20} \cmidrule(l{2pt}r{2pt}){21-22} \cmidrule(l{2pt}r{2pt}){24-25}
&&& Acc↑ & \#\,Fwd↓ && Acc↑ & \#\,Fwd↓ & Acc↑ & \#\,Fwd↓  & &Acc↑ &\#\,Fwd↓ &&& EM↑ & \#\,Fwd↓ && EM↑ & \#\,Fwd↓ & EM↑ & \#\,Fwd↓  & & EM↑ &\#\,Fwd↓ \\
\midrule
\multicolumn{25}{c}{\textcolor{gray!90}{\textbf{\textit{Qwen2.5-0.5B-Instruct}}}}\\
\midrule
Explicit-CoT & \xmark  && 50.9 & 117.3 && 13.4 & 160.0 & 59.0 & 61.9  & & 41.1&113.1&&& 13.2 & 951.2 && 21.0 & 739.1 & 11.3 & 891.6 & & 15.2&860.6\\
TokenSkip           & \xmark && 46.2 & 73.6 && 10.4 & 104.7 & 49.8 & 38.5  & & 35.5&72.3&&& 14.2 & 502.8 && 21.7 & 452.3 & 9.9 & 474.1 & & 15.3 & 476.4\\
PCCoT               & \xmark && 20.6 & 8.4 && 4.9 & 10.7 & 29.1 & 8.0  & & 18.2&9.0&&& 11.2 & 9.7 && 20.6 & 10.5 & 2.8 & 9.4  & & 11.5&9.9\\
CODI                & \xmark && 34.2 & 14.5 && 9.0 & 17.0 & 42.8 & 14.0  & & 28.7&15.2&&& 13.0 & 13.6 & & 20.7 & 14.3 & 5.9 & 13.1  & & 13.2&13.7\\
SIM-CoT             & \xmark && 39.3 & 14.5 && 8.8 & 17.5 & 45.0 & 14.1  & & 31.0&15.4&&& 13.0 & 13.6 & & \textbf{22.0} & 14.4 & 5.7 & 13.1 & & 13.6&13.7\\
\rowcolor{gray!15}  \algo & \xmark  && \textbf{58.8} & 54.6 && \textbf{14.2} & 75.2 & \textbf{63.3} & 31.3  & & \textbf{45.4} &53.7&&& \textbf{14.2} & 615.1 &  & {21.8} & 497.8 & \textbf{11.8} & 660.4  & & \textbf{15.9}&591.1\\
\noalign{\vskip 2pt}
\cdashline{1-25}
\noalign{\vskip 2pt}
 Explicit-CoT & \cmark && {52.4} & 109.2 && {30.4} & 145.7 & {62.4} & 58.4  & & 48.4&104.4&&& {33.1} & 309.0 && {40.8} & 247.8 & {53.3} & 293.0  & & 42.4&283.3\\
\rowcolor{gray!15} \algo
                                & \cmark && \textbf{61.2}* & 43.9 && \textbf{32.8} & 57.0 & \textbf{67.7}$^\dagger$ & 26.4  & & \textbf{53.9}&42.4&&& \textbf{36.2}$^\dagger$ & 49.4 && \textbf{46.5}* & 39.8 & \textbf{54.8} & 43.3 & & \textbf{45.8}&44.2\\
\midrule
\multicolumn{25}{c}{\textcolor{gray!90}{\textbf{\textit{Llama-3.2-1B-Instruct}}}}\\
\midrule
Explicit-CoT & \xmark  && 55.1 & 97.0 && 12.4 & 112.9 & 62.3 & 50.5  & & 43.2&86.8&&& 17.4 & 1126.1 && 19.8 & 928.5 & \textbf{30.9} & 957.6 && 22.7&1004.1\\
TokenSkip           & \xmark && 52.8 & 57.2 && 11.6 & 67.9 & 60.3 & 28.1  && 41.6 & 51.1 &&& 17.1 & 438.8 && 22.0 & 361.3 & 24.0 & 387.1  &&  21.0 & 395.7\\
PCCoT               & \xmark && 45.6 & 7.5 && 10.8 & 7.9 & 53.9 & 7.5  && 36.8 & 7.6 &&& 4.6 & 11.3 && 7.9 & 11.9 & 4.8 & 10.6  && 5.8 & 11.3\\
CODI                & \xmark && 55.4 & 15.0 && 12.9 & 17.2 & 61.3 & 15.4  && 43.2 & 15.9&&& 18.8 & 13.4 && 22.9 & 14.1 & 24.8 & 13.0  & & 22.2&13.5\\
SIM-CoT             & \xmark && 56.0 & 15.0 && 12.6 & 20.5 & 61.2 & 16.5  && 43.3 & 17.3 &&& \textbf{19.1} & 13.4&& \textbf{23.8} & 14.1 & 26.5 & 12.9  & & 23.1&13.5\\
\rowcolor{gray!15}
\algo & \xmark  && \textbf{61.9} & 45.3 && \textbf{14.3} & 52.6 & \textbf{69.0} & 26.9  & & \textbf{48.4} &41.6&&& 18.3 & 592.7 && 22.8 & 554.6 & 28.8 & 512.4  & & \textbf{23.3} & 553.2\\
\noalign{\vskip 2pt}
\cdashline{1-25}
\noalign{\vskip 2pt}
 Explicit-CoT & \cmark && {58.2} & 86.1 && {35.7} & 97.3 & {66.2} & 45.4  & & 53.4&76.3&&& {39.9} & 270.2 && {48.2} & 225.1 & {65.1} & 240.5  & & 51.1&245.3\\
\rowcolor{gray!15} \algo 
                                & \cmark          && \textbf{65.1}* & 38.2 && \textbf{39.3}* & 43.5 & \textbf{73.2}* & 23.6  & & \textbf{59.2}&35.1&&& \textbf{41.4} & 52.8 && \textbf{52.0}* & 44.7 & \textbf{65.8} & 46.6 & & \textbf{53.1}&48.0\\

\midrule
\multicolumn{25}{c}{\textcolor{gray!90}{\textbf{\textit{Qwen3-4B}}}}\\
\midrule
Explicit-CoT & \xmark  && 76.7 & 133.2 && 44.1 & 195.7 & 77.4 & 71.9 && {66.1} & 133.6 &&& 11.2 & 1496.0 && 10.6 & 1524.5 & 23.9 & 1273.7 && 15.2 & 1431.4 \\
\rowcolor{gray!15} \algo & \xmark  && \textbf{82.6} & 52.6 && \textbf{45.9} & 77.3 & \textbf{85.5} & 28.3 && \textbf{71.3} & 52.7 &&& \textbf{20.4} & 400.1 && \textbf{26.3} & 381.4 & \textbf{46.9} & 328.6 && \textbf{31.2} & 370.0\\
\noalign{\vskip 2pt}
\cdashline{1-25}
\noalign{\vskip 2pt}
Explicit-CoT & \cmark && 78.0 & 115.2 && 49.1 & 154.3 & {78.5} & 61.7 && 68.5 & 110.4 &&& {43.4} & 272.2 && {59.9} & 224.5 & {72.3} & 235.3 && 58.5 & 244.0\\

\rowcolor{gray!15} \algo & \cmark && \textbf{83.4}* & 41.6 && \textbf{54.2}* & 55.9 & \textbf{86.0}* & 23.8 && \textbf{74.5} & \textbf{40.4} &&& \textbf{45.9}* & 43.1 && \textbf{60.5} & 44.9 & \textbf{75.5}* & 37.4 && \textbf{60.6} & \textbf{41.8}\\

\bottomrule
\end{tabular}}
\vspace*{-0.2cm}
\caption{Main results on the math and QA benchmarks ({Qwen2.5-0.5B, Llama-3.2-1B, and Qwen3-4B}). \#\,Fwd is the average number of forward passes per query. \textbf{TU} indicates whether tools are used. Best scores are highlighted in \textbf{bold};
* and $\dagger$ mark $p<0.01$ and $p<0.05$ improvements over the second-best via paired bootstrap with $10^5$ samples~\citep{koehn-2004-statistical}.
}
\vspace*{-0.3cm}
\label{tab:main}
\end{table*}

\section{Experiments}

We design experiments to evaluate two capabilities of our proposed method:  (1) agentic math problem-solving training and evaluation to assess its reasoning capability, and (2) agentic QA training and evaluation to assess its agentic capability.

\subsection{Experimental Setup}


\paragraph{Training Datasets.}
%
\algo requires training to use implicit tokens. To evaluate mathematical reasoning, we use \textbf{GSM8K-Aug-NL}~\citep{deng2024implicit}, which contains explicit interactions as textual explanations. Since GSM8K-Aug-NL contains explicit reasoning trajectories without executable interfaces, we insert executable calculator annotations from GSM8K-Aug~\citep{deng2024implicit}. To evaluate QA capability, we use the \textbf{HotpotQA}~\cite{yang-etal-2018-hotpotqa} dataset. We construct 40{,}000 trajectories by using DeepSeek-V3~\citep{deepseekai2025deepseekv3technicalreport} with the available tool and allowing it to solve each question through iterative tool interaction. 
{
We retain only trajectories whose final answer matches the gold annotation.
Although teacher-generated trajectories may inherit teacher-specific biases, TAC-CoT and all QA baselines are trained on the same trajectories; such biases are therefore shared across methods and cannot by themselves explain TAC-CoT's relative improvements.
Dataset construction details are provided in Appendix~\ref{sec:appendix-data}.
}

\paragraph{Evaluation Benchmarks.}
We evaluate \algo on agentic math problem solving (Math), agentic Question-Answering (QA), and tool-use benchmarks, as summarized in Table~\ref{tab:evaluation_benchmarks}. On toolBench, we use `Single' for the I1 single-tool scenario, and `Multi' as the I2\,+\,I3 multi-tool scenario.
We report accuracy (Acc) for math tasks and exact match (EM) for QA. 
To compare the inference cost 
{
independently of hardware and external API latency
}, we report the average number of forward passes (\#\,Fwd)~\cite{kuzina2026kava, wei2026simcot}. 
{
For no-tool QA setting (TU = \xmark), \# Fwd also includes autoregressive generation of missing tool observations; see Appendix~\ref{sec:implicitcot_implementation_details}.
}

\paragraph{Implementation Details.} 
We use Qwen2.5-0.5B-Instruct~\citep{qwen2.5}, Llama-3.2-1B-Instruct~\citep{grattafiori2024llama}, and Qwen3-4B~\cite{yang2025qwen3technicalreport} as backbone models. 
For agentic QA, we use the Wikipedia API as the search engine, following \citet{yao2022react}. Further details are provided in Appendix~\ref{sec:implementation_details}. We provide pseudocode for reproducibility in Figure~\ref{fig:inference-pseudocode}. For \algo, we set $B=3$. Ablation results for different $B$ values are provided in Appendix~\ref{sec:continuous_token}.



\paragraph{Baselines.} 
We compare \algo against (1) an \textbf{Explicit-CoT} agent, which serves as a strong baseline for tool use and reasoning by using full explicit reasoning traces, fine-tuned on the same training trajectories as our method; (2) \textbf{TokenSkip}~\cite{xia-etal-2025-tokenskip}, an explicit CoT method that compresses intermediate reasoning trajectories by pruning low-importance tokens; (3) \textbf{PCCoT}~\cite{wu-etal-2025-parallel}, which updates implicit thought tokens in parallel using Jacobi iteration, improving efficiency over sequential implicit CoT; (4) \textbf{CODI}~\cite{shen-etal-2025-codi}, an implicit-CoT method trained via trajectory-level self-distillation from CoT-based teacher signals; and (5) \textbf{SIM-CoT}~\cite{wei2026simcot}, which extends CODI by adding step-level supervision to reconstruct reasoning steps from latent embeddings.

\subsection{Main Results}\label{sec:main_results}

Our evaluation examines whether \algo achieves its design goals in addressing the limitations of extending existing CoT approaches to agentic systems, focusing on four aspects: tool-use effectiveness, internal reasoning ability, query-adaptive continuous-thought budgeting, and inference efficiency. Table~\ref{tab:main} presents the main results on agentic math and QA benchmarks across tool-calling and no-tool settings.

\paragraph{Performance on Agentic Reasoning.}
Overall, \algo consistently outperforms Explicit-CoT and the other baselines. Averaged over all backbone models, Our method improves math accuracy by 5.8\%p and QA performance by 2.4 points compared with Explicit-CoT. The gains also transfer to out-of-domain benchmarks: Ours improves math performance by 6.5\%p in-domain and 5.5\%p out-of-domain, and QA performance by 2.4 points in-domain and 2.5 points out-of-domain. This consistent trend suggests that \algo provides robust adaptive implicit reasoning beyond the training-domain distribution.

To isolate the effect of implicit reasoning from external tool execution, we compare \algo with other baselines in the no-tool setting on math datasets.
While TokenSkip, which shortens rationales by pruning tokens, underperforms Explicit-CoT, \algo consistently outperforms the Explicit-CoT agent across all math datasets.


{
These results indicate that the improvement in internal reasoning does not simply come from shorter reasoning traces.
Consistently, TAC-CoT also outperforms baselines that either remove natural-language reasoning entirely or retain concise, explicit reasoning with tool use; see Appendix~\ref{app:controlled_baselines}.
}


\paragraph{Effectiveness of \algo.}
The effectiveness of \algo comes from both tool-use and query-adaptive implicit reasoning.
Tool-use substantially improves QA performance, yielding up to a 29.9 point average gain over the no-tool variant while reducing forward passes by 13.3\(\times\) on Qwen2.5-0.5B. Even without tools, \algo benefits from adaptive budgeting: on math benchmarks, its no-tool variant outperforms prior implicit-CoT baselines while requiring far fewer forward passes than explicit CoT. These results show that \algo improves efficiency and performance by adaptively allocating continuous reasoning, with additional gains when external knowledge is needed. 

Supplementary full fine-tuning results are provided in Appendix \ref{sec:finetuning}.

\subsection{Diagnostic Analysis} \label{sec:advantage}
We conduct diagnostic analyses to identify the sources of \algo's performance gains across four dimensions: tool-use quality, reduced error accumulation, query-adaptive budgeting efficiency, and lower inference-token cost.

\begin{table}[t]
\renewcommand{\arraystretch}{1.0}
\centering
\setlength{\tabcolsep}{5.5pt}
\resizebox{0.48\textwidth}{!}{

\begin{tabular}{l | cc cc cc}
\toprule
\multirow{2.5}{*}{\textbf{Method}} 
& \multicolumn{2}{c}{\textbf{HotpotQA}}
& \multicolumn{2}{c}{\textbf{2WikiMulti}}
& \multicolumn{2}{c}{\textbf{TriviaQA}}\\
\cmidrule(l{2pt}r{2pt}){2-3} \cmidrule(l{2pt}r{2pt}){4-5} \cmidrule(l{2pt}r{2pt}){6-7}
& @5 & @10 & @5 & @10& @5 & @10\\
\midrule
Explicit-CoT        & 48.8 & 51.7 & 70.9 & 71.6 & 72.6 & 75.5 \\
\rowcolor{gray!15}  \algo & \textbf{51.8}$^\dagger$ & \textbf{55.4}* & \textbf{73.7}* & \textbf{74.4}$^\dagger$ & \textbf{74.5} & \textbf{77.4} \\
\bottomrule
\end{tabular}}

\vspace*{-0.2cm}
\caption{Answer Coverage @ $k$ (\%) over retrieval steps on the QA benchmarks, Qwen2.5-0.5B-Instruct. All AC values are reported as percentages; Best scores are \textbf{bold}; * and $\dagger$ mark $p<0.01$ and $p<0.05$ improvements over the underlined Explicit-CoT.
\looseness=-1
}

\vspace*{-0.1cm}
\label{tab:action_decision}
\end{table}

\begin{figure}[!t]
    \centerline{\includegraphics[width=0.97\columnwidth]{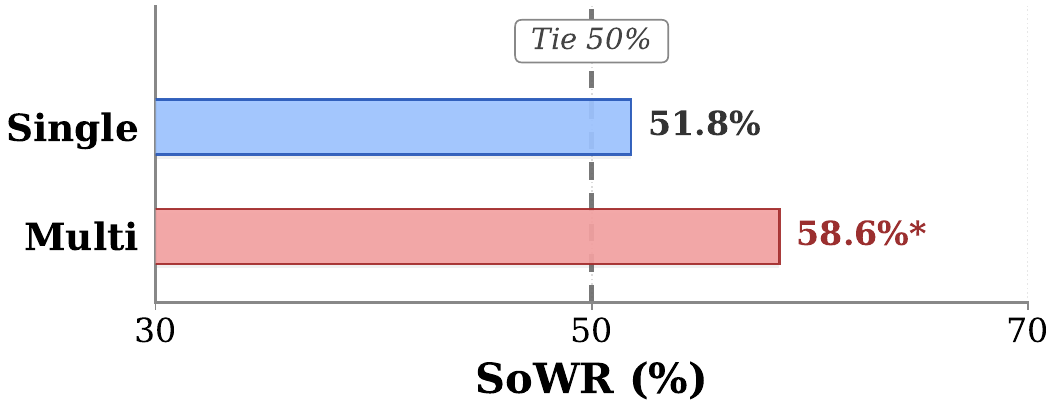}}
    \vspace*{-0.2cm}
    \caption{Solvable win rate (SoWR) of \algo against Explicit-CoT on ToolBench, judged by GPT-5.1. `Single' denotes the single-tool scenario, and `Multi' denotes the multi-tool scenario. * marks $p<0.01$.
    \looseness=-1
    }
    \vspace*{-0.3cm}
    \label{fig:toolbench}
\end{figure}

\paragraph{Effectiveness of Tool Use.}
We evaluate whether \algo improves tool-mediated decisions on QA benchmarks using Answer Coverage@$k$ (AC@$k$), which measures whether the gold answer appears in the first \(k\) retrieved observations. As shown in Table~\ref{tab:action_decision}, \algo improves AC@5 and AC@10 across all QA datasets, with significant gains on HotpotQA and 2WikiMultihopQA. Notably, its AC@5 is comparable to the AC@10 of Explicit-CoT, indicating that \algo retrieves supporting evidence in fewer steps.
We further analyze tool-use complexity on ToolBench by fine-tuning Qwen2.5-0.5B-Instruct and comparing \algo with Explicit-CoT under the StableToolBench evaluation protocol~\cite{guo-etal-2024-stabletoolbench}. As shown in Figure~\ref{fig:toolbench}, \algo achieves Solvable Win Rates of 51.8\% on the single-tool and 58.6\% in the multi-tool scenario, with the latter gain being statistically significant. This suggests that interleaved implicit reasoning is especially beneficial for complex multi-tool interactions. A detailed analysis is provided in Appendix~\ref{sec:tool_use_budget}.


\begin{table}[t]
\setlength{\tabcolsep}{6.5pt}
\renewcommand{\arraystretch}{1.0}
\centering
\resizebox{0.48\textwidth}{!}{
\begin{tabular}{l|c|cccc|c}
\toprule
\multirow{2.5}{*}{\textbf{Metric}}
& \multirow{2.5}{*}{\textbf{E-CoT}}
& \multicolumn{4}{c|}{\textbf{\# Replaced}}
& \multirow{2.5}{*}{\textbf{Ours}} \\
\cmidrule(lr){3-6}
& & {1} & {2} & {3} & {4} & \\
\midrule
Mean Acc\,$\uparrow$
& 41.1
& 41.7
& 42.4
& 43.4
& 44.9
& \cellcolor{gray!15} \textbf{45.4} \\
\#\,Fwd\,$\downarrow$
& 113.1
& 96.2
& 79.2
& 63.4
& 58.9
& \cellcolor{gray!15} \textbf{53.7} \\
\bottomrule
\end{tabular}}
\vspace*{-0.15cm}
\caption{Effect of replacement depth on Qwen2.5-0.5B-Instruct average accuracy over GSM8K, GSM-Hard, and SVAMP. All accuracy values are reported as percentages. 
\textbf{E-CoT} denotes the Explicit-CoT agent, and \emph{\# Replaced} denotes the number of explicit reasoning segments swapped for continuous thought tokens.
}
\vspace*{-0.2cm}
\label{tab:compression}
\end{table}

\begin{figure}[!t]
    \centerline{\includegraphics[width=0.97\columnwidth]{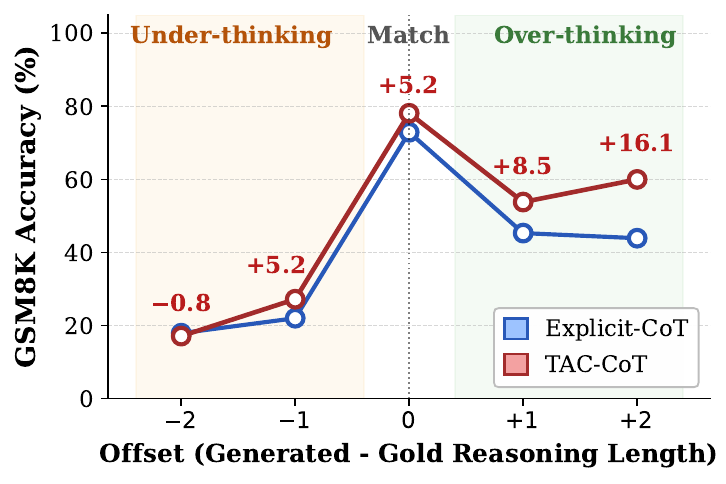}}
\vspace*{-0.25cm}
    \caption{GSM8K accuracy stratified by the offset between the number of generated reasoning steps and the gold step count in the gold solution (negative $=$ under-thinking, $0$ $=$ matched depth, positive $=$ over-thinking).}
    \label{fig:step_calibration}
    \vspace*{-0.05cm}
\end{figure}

\paragraph{Error Accumulation Mitigation.}
Table~\ref{tab:main} shows that \algo improves internal reasoning even without external execution. To test whether this comes from reduced error accumulation, we analyze reasoning compression and length mismatch.   
First, we progressively replace explicit reasoning segments with implicit ones when training Qwen2.5-0.5B-Instruct on the augmented GSM8K trajectories. 
We vary the replacement depth from one to four segments, since four corresponds to the 90th percentile of the number of explicit reasoning segments in the training trajectories.
As shown in Table~\ref{tab:compression}, average accuracy on GSM8K, GSM-Hard, and SVAMP increases monotonically as more segments become implicit, suggesting that implicit CoT mitigates errors from lengthy rationales.
Second, following the step-efficiency analysis of prior work~\cite{wu2026when}, we group GSM8K examples by the offset between generated and gold reasoning lengths. Figure~\ref{fig:step_calibration} shows that both Explicit-CoT and \algo perform best under matched lengths and degrade under over-thinking offsets. However, \algo degrades less, indicating that implicit reasoning reduces the penalty from unnecessary intermediate steps. 
We discuss further analysis of error accumulation in Appendix~\ref{sec:error_accumulation}.

\begin{figure}[!t]
    \centerline{\includegraphics[width=0.97\columnwidth]{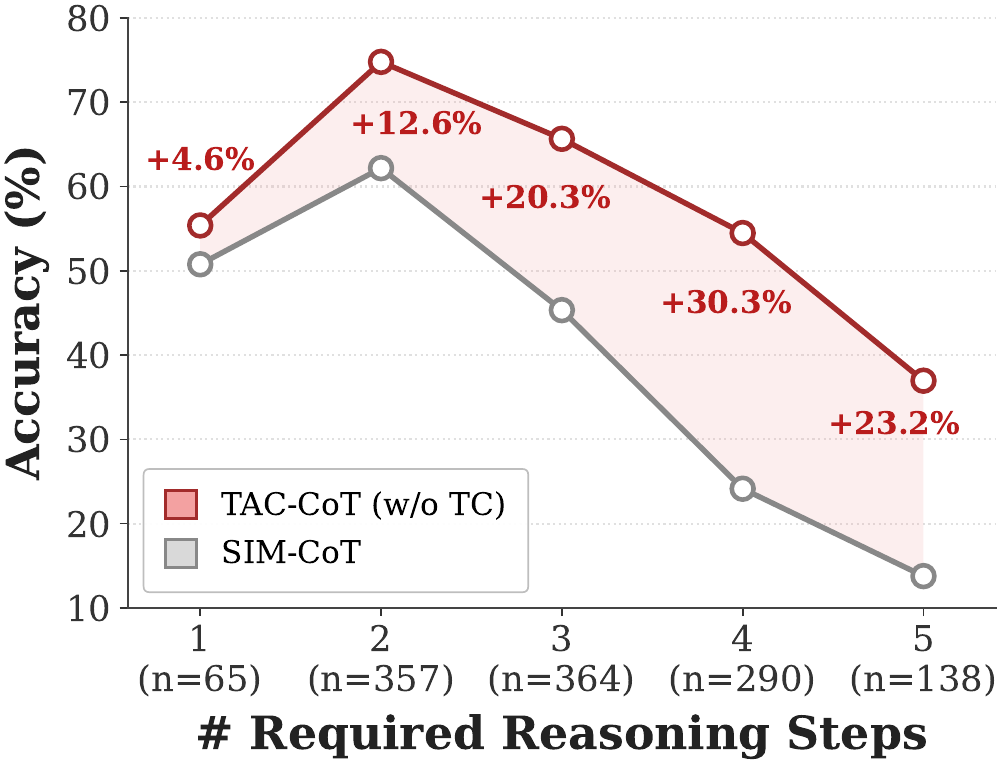}}
    \vspace*{-0.2cm}
    \caption{{Accuracy comparison across GSM8K subsets grouped by the number of required reasoning steps. 
    Each $n$ denotes the number of samples in the corresponding reasoning-step group.}}
    \label{fig:querydependent}
    \vspace*{-0.2cm}
\end{figure}

\paragraph{Query-Adaptive Continuous-Thought Budget.}
Unlike prior implicit-CoT methods that use a fixed continuous-token budget for every query, \algo adaptively allocates its total reasoning budget according to the query. 
This query-adaptive continuous-thought budgeting contributes to the performance gains, especially for math tasks that depend on internal reasoning. 
As shown in Figure~\ref{fig:querydependent}, the benefit becomes larger as the required number of reasoning steps increases.

\algo's query-adaptive budgeting benefits inputs that require different amounts of internal deliberation before producing the final answer. 
We group GSM8K examples into five subsets according to the number of gold reasoning steps, ranging from one to five steps, and treat this number as the required reasoning depth. 
We then evaluate each subset separately and compare \algo without tool use against SIM-CoT, a state-of-the-art fixed-budget implicit-CoT baseline, using Qwen2.5-0.5B-Instruct. 
Figure~\ref{fig:querydependent} shows that \algo consistently outperforms SIM-CoT across all reasoning-depth groups, with larger gains on examples that require more reasoning steps. 
The improvement increases from 4.6 points on one-step examples to 30.3 points on four-step examples.
This trend suggests that assigning the same continuous-thought-token budget to every input becomes increasingly restrictive as the required number of reasoning steps grows. 
In contrast, \algo increases the total continuous-token budget by continuing through additional implicit reasoning segments when more reasoning steps are needed. 
These results demonstrate that \algo flexibly allocates reasoning budget across queries, which is especially important for reasoning-intensive tasks.

\paragraph{Inference Efficiency.} \label{sec:inference_cost}
\algo substantially reduces inference cost relative to Explicit-CoT while improving task accuracy. 
Across backbones, it reduces the per-query forward count by up to $2.6\times$ on math benchmarks and $6.8\times$ on QA benchmarks, with larger savings on QA due to longer multi-step trajectories. 
This suggests that \algo becomes increasingly beneficial for longer trajectories.

Although fixed-budget implicit-CoT baselines use fewer forward passes, they do not adapt to query complexity or retrieve missing observations. 
This limitation is especially harmful for QA, where answers often depend on retrieved evidence, leading to pronounced performance degradation. 
In contrast, \algo achieves a better cost–performance trade-off by reducing inference cost relative to Explicit-CoT while avoiding the QA degradation of fixed-budget implicit-CoT methods.

{On a single NVIDIA L40S GPU, \algo achieves measured wall-clock speedups of 2.3× on math and 2.4× on QA. Since realized latency depends on hardware, implementation, and external API overhead, \#Fwd reductions should not be interpreted directly as wall-clock speedups; see Appendix~\ref{sec:latency} for full results.}

\section{Related Works}

\paragraph{Explicit and Implicit CoT.}
Chain-of-Thought (CoT) improves the reasoning ability of large language models by generating intermediate reasoning before the final answer~\cite{wei2022chain,kojima2022large}. 
CoT methods are commonly categorized as explicit or implicit, depending on whether intermediate reasoning is represented as discrete tokens or continuous representations. 
Explicit CoT is readable and easy to supervise~\cite{wei2022chain,kojima2022large,yao2023tree}, but decoding long reasoning traces increases inference cost~\cite{han-etal-2025-token,xia-etal-2025-tokenskip} and exacerbates stepwise error accumulation~\cite{wu2026when}. 
\looseness=-1

Implicit CoT instead performs intermediate reasoning in continuous latent space~\cite{hao2025training,shen-etal-2025-codi,wu-etal-2025-parallel,wei2026simcot,kuzina2026kava}. 
For example, Coconut feeds hidden states back as continuous thought tokens~\cite{hao2025training}, CODI distills explicit CoT into continuous representations~\cite{shen-etal-2025-codi}, PCCoT updates latent thought tokens in parallel for efficient inference~\cite{wu-etal-2025-parallel}, and SIM-CoT stabilizes implicit reasoning via step-level reconstruction~\cite{wei2026simcot}. 
Although these methods reduce the cost of verbalized reasoning, they are mainly designed for standalone reasoning: they typically allocate a fixed number of continuous thought tokens per query and do not expose intermediate decisions as executable interfaces, limiting their suitability for agentic settings that require repeated reasoning and acting.

{
In contrast, TAC-CoT extends implicit CoT to tool-using agentic systems by interleaving compact latent reasoning segments with explicit tool interactions, enabling tool calling and query-adaptive reasoning budgets.
At the same time, compared with explicit-CoT agents, TAC-CoT avoids the need to verbalize intermediate reasoning, thereby reducing inference cost and mitigating error accumulation from lengthy reasoning traces.
}

\paragraph{Tool-Using LLM Agents.}
LLMs are increasingly studied as agents that solve tasks through interleaved reasoning, tool calls, and observations~\cite{yao2022react,schick2023toolformer,shinn2023reflexion,yang2025qwen3technicalreport}. 
ReAct formalizes this reasoning--acting loop~\cite{yao2022react}, while Toolformer~\cite{schick2023toolformer} and ART~\cite{paranjape2023artautomaticmultistepreasoning} show that models can invoke external APIs and resume generation after tool execution. 
However, most tool-using agents still rely on explicit CoT~\cite{wei2022chain,kojima2022large,yao2023tree}, whose repeated generation across multi-step interaction loops makes inference costly~\cite{han-etal-2025-token,xia-etal-2025-tokenskip,aytes-etal-2025-sketch} and vulnerable to error accumulation~\cite{wu2026when}. 

\section{Conclusion}
\label{sec:conclusion}

We presented \algo, a tool-empowered adaptive continuous CoT framework that extends implicit reasoning to tool-using agentic problem solving. By interleaving compact continuous reasoning segments with explicit tokens for tool calls, observations, and final answers, \algo preserves the executable interface required for agentic workflows while avoiding the cost of fully verbalized reasoning traces. Across agentic math and knowledge-intensive QA benchmarks, \algo consistently outperforms Explicit-CoT agents in both in-domain and out-of-domain settings, while using substantially fewer model forward passes. Further analyses show that \algo improves tool-use decisions, performs better in multi-tool scenarios, mitigates error accumulation, and benefits from query-adaptive continuous thought-token allocation. 
\looseness=-1

\section*{Limitations}
This work has a few limitations:
First, the adaptive routing mechanism operates at the segment level rather than the token level. Second, longer interaction horizons remain an important direction for evaluating the practical deployment of \algo{}.

\section*{Acknowledgements}
This work was supported by the National Research Foundation of Korea (NRF) grant funded by the Korea government (MSIT) (No. RS-2024-00334343).
Additionally, this work was supported by the National IT Industry Promotion Agency (NIPA) grant funded by the Korea government (MSIT) (No. RS-2026-25621604).

\bibliography{anthology}
\bibliographystyle{acl_natbib}

\appendix
\clearpage

\section{Details of Data Construction}
\label{sec:appendix-data}
This section provides the details of the recipe described in Section~\ref{sec:data} for each domain.

\paragraph{Math Problem Solving.} 
The math domain follows the source-obtainable interaction setting described.
For GSM8K, where data with explicit interactions already exist, we use GSM8K-Aug and GSM8K-Aug-NL, two GSM8K augmentations with 385k samples each~\cite{deng2024implicit}. GSM8K-Aug retains only the arithmetic equations in each reasoning step (e.g., \texttt{<<86.7*15*.01=13.005>>}), whereas GSM8K-Aug-NL contains only explicit reasoning (e.g. \texttt{...day, 86.7 x 15\% = 13.005 coins...}). Because the two datasets have the same questions, we align examples by question and insert the equations from GSM8K-Aug into the corresponding natural language reasoning trajectory from GSM8K-Aug-NL.

\paragraph{Question-Answering.}
\label{sec:hotpot_distill}
For HotpotQA, which provides only questions and their gold answers without faithful tool-use trajectories, we construct trajectories by prompting DeepSeek-V3 to solve each question with access to a retrieval tool.
Following \citet{yao2022react}, we implement a Wikipedia retrieval system and use the ReAct-style few-shot demonstrations from their setup.
The prompt template is shown in Figures~\ref{fig:qa_prompt_1} and~\ref{fig:qa_prompt_2}.


{
\paragraph{Quality Audit and Construction Cost.}
To assess the quality of the synthesized QA trajectories, we randomly sampled 200 trajectories and evaluated them using GPT-5.1, a model different from the DeepSeek-V3 teacher.
The evaluator assessed whether the generated actions and final answers were logically valid, and judged 99\% of the sampled trajectories to be valid.
}

{
Generating the QA trajectories cost \$102.73 in total, including generations discarded during answer-based filtering.
The filtering pass rate was 56.24\%.
Because exact-match filtering may reject semantically correct answers expressed differently from the gold annotation, this procedure may underestimate the usable trajectory yield.
}

\section{Implementation Details}
\label{sec:implementation_details}

\subsection{Training hyperparameters}
For the CODI loss weight, we set $\alpha_{\mathrm{CODI}}=1$ for \algo and $\alpha_{\mathrm{CODI}}=10$ for PCCoT, CODI, and SIM-CoT, following \citet{shen-etal-2025-codi, wu-etal-2025-parallel, wei2026simcot}.
For TokenSkip, we use a 50\% compression ratio during training to keep its number of forward passes comparable to that of \algo.
{For the larger Qwen3-4B backbone, we use $B=2$ continuous tokens per segment for a practical training setup.}

\subsection{Implicit CoT Baseline Details}
\label{sec:implicitcot_implementation_details}
For each implicit CoT baseline, we report the higher score between models trained on GSM8K-Aug-NL and GSM8K-Aug.

\paragraph{Evaluation benchmark.}
{For math benchmarks, we evaluate on the full test sets. 
For QA benchmarks, we evaluate on 1,000 randomly sampled examples from each dataset.
This follows approaches used by \citet{yin-etal-2023-exchange,kaesberg-etal-2025-voting,jang-etal-2025-dice}.
Specifically, we follow the standard sample-size estimate for population proportions:
\[
n_0 = \frac{Z^2 \cdot p \cdot (1-p)}{d^2},
\]
where \(Z=1.96\), \(p=0.5\), and \(d=0.05\), corresponding to a 95\% confidence level and a 5\% margin of error under the conservative maximum-variance assumption. 
This gives \(n_0 \approx 384\), so our choice of 1,000 examples per QA dataset is above the required sample size. 
Since EM and Answer Coverage are averaged binary outcomes over examples, 1,000 random samples yield a conservative worst-case 95\% margin of error of approximately \(1.96\sqrt{0.25/1000}=3.1\) percentage points before finite-population correction.}


\paragraph{Calculator setting.}
For math tasks, we build a calculator database by extracting equations from the training trajectories. During evaluation, if an expression matches an entry in this database, we use the corresponding stored result. Otherwise, we process the expression before calculation and then execute it with the calculator. This design handles cases where expressions contain units or other non-numeric tokens that cannot be directly evaluated by a standard calculator.

\paragraph{ToolBench API details.} For ToolBench, many of the original APIs are no longer available at inference time. Following \citet{guo-etal-2024-stabletoolbench}, we therefore use MirrorAPI. Because \algo does not verbalize its reasoning, we explicitly note this in the GPT-5.1 judge prompt.

\begin{table}[t]
\centering
\setlength{\tabcolsep}{2pt}
\resizebox{0.5\textwidth}{!}{
\begin{tabular}{lccc}
\toprule
\textbf{Hyperparameter} 
& \shortstack{\textbf{Qwen2.5}\\\textbf{0.5B-Instruct}} 
& \shortstack{\textbf{Llama-3.2}\\\textbf{1B-Instruct}} 
& \shortstack{\textbf{Qwen3}\\\textbf{4B}} \\
\midrule
Fine-tuning strategy & Full fine-tuning & LoRA & LoRA\\
LoRA rank            & --              & 128 & 128\\
LoRA scaling factor  & --              & 32 & 32\\
LoRA dropout         & --              & 0.1 & 0.1\\
Learning rate        & $1\times10^{-4}$ & $8\times10^{-4}$ & $1\times10^{-4}$ \\
LR scheduler         & Cosine          & Cosine & Cosine \\
Warmup               & 0.03            & 0.03 & 0.03 \\
Optimizer            & AdamW           & AdamW & AdamW \\
Batch size           & 128             & 128 & 128 \\
Gradient norm clipping & 2             & 2 & 2 \\
Training epochs      & 10              & 10 & 5 \\
\bottomrule
\end{tabular}}
\caption{
Training hyperparameters used for each backbone model. Values are shared across all compared methods unless otherwise specified.
}
\label{tab:hyperparam}
\end{table}

\paragraph{No-tool evaluation protocol.}
For the no-tool setting (TU = \xmark), we disable the external tool executor while preserving the trajectory format used by tool-using agents. When the model emits an action-like span, no environment-returned observation is injected. Instead, the model continues generation and produces the corresponding observation text by itself. Thus, the no-tool setting evaluates reasoning without externally supplied tool outputs. All generated observation tokens are counted in the inference cost, and we use the same answer extraction and evaluation metrics as in the tool-use setting.

\section{Additional Experiments}
\label{sec:appendix}
This appendix provides additional analyses that complement the main experimental results. We first study how the number of allowed tool calls can mitigate error propagation~\cite{yao2022react}. We then analyze how errors accumulate as reasoning depth increases, which motivates the need for shorter reasoning methods, such as implicit CoT. Finally, we report supplementary full fine-tuning results for additional backbone settings that are not included in the main results table.





\begin{table}[t]
\renewcommand{\arraystretch}{1.0}
\centering
\setlength{\tabcolsep}{7pt}
\resizebox{0.47\textwidth}{!}{
\begin{tabular}{c|ccc|c}
\toprule
\multirow{2.5}{*}{\textbf{\#\,CT}}
& \multicolumn{4}{c}{\textbf{Math}} \\
\cmidrule(l{2pt}r{2pt}){2-5}
& {GSM8K} & {GSM-Hard} & {SVAMP} & {Average} \\
\midrule
0 & 55.7 & 13.9 & 56.3 & 42.0 \\
1 & 58.5 & \textbf{14.7} & 61.5 & 44.9 \\
3 & \textbf{58.8} & 14.2 & \textbf{63.3} & \textbf{45.4} \\
\bottomrule
\end{tabular}}
\caption{Effect of the number of continuous thought tokens (\#\,CT) per continuous segment on math datasets, using Qwen2.5-0.5B-Instruct. All scores are reported as percentages.}
\label{tab:num_con_token}
\end{table}

\subsection{Effect of Continuous-Thought Tokens}\label{sec:continuous_token}

We analyze the effect of the number of continuous thought tokens (CT) per continuous segment. We vary the CT budget over $\{0,1,3\}$ and evaluate Qwen2.5-0.5B-Instruct on GSM8K, GSM-Hard, and SVAMP. Table~\ref{tab:num_con_token} shows the results. Compared to the 0 CT setting, using CT tokens improves the average accuracy from $42.0\%$ to $44.9\%$ with one CT token and $45.4\%$ with three CT tokens. These results suggest that allocating more CT can be beneficial.

\subsection{Controlled Baseline Comparisons}
\label{app:controlled_baselines}

{
We conduct additional controlled comparisons to distinguish the benefit of implicit reasoning from two simpler explanations: removing natural-language rationales and shortening explicit reasoning traces.
All experiments in this section use Qwen2.5-0.5B-Instruct.
}

{
\paragraph{No Reasoning.}
We first consider a no reasoning control.
This baseline removes natural-language rationales, while retaining the arithmetic equations used for intermediate computation.
This comparison isolates whether TAC-CoT's improvement can be explained simply by removing verbose natural-language reasoning.
As shown in Table~\ref{tab:controlled_baselines}, TAC-CoT improves the average accuracy from 42.0\% to 45.4\% in the no-tool setting.
}

{
\paragraph{Concise Explicit-CoT with Tools.}
We further compare against a concise explicit-CoT agent with the same tool-enabled setting.
Specifically, we extend TokenSkip to tool use and apply the same 50\% compression ratio used in our main experiments, thereby shortening explicit reasoning traces while preserving explicit reasoning and tool interaction.
TAC-CoT achieves an average accuracy of 53.9\%, compared with 45.4\% for TokenSkip with tools.
The improvement is consistent across GSM8K, GSM-Hard, and SVAMP.
}

\begin{table}[t]
\renewcommand{\arraystretch}{1.0}
\centering
\setlength{\tabcolsep}{7pt}
\resizebox{0.47\textwidth}{!}{
\begin{tabular}{lcccc}
\toprule
Method & GSM8K & GSM-Hard & SVAMP & Average \\
\midrule
\multicolumn{5}{l}{\textit{No-tool setting}} \\
No Reasoning & 55.7 & 13.9 & 56.3 & 42.0 \\
\rowcolor{gray!15}
\textbf{TAC-CoT} & \textbf{58.8} & \textbf{14.2} & \textbf{63.3} & \textbf{45.4} \\
\midrule
\multicolumn{5}{l}{\textit{Tool-enabled setting}} \\
TokenSkip & 52.3 & 27.0 & 57.0 & 45.4 \\
\rowcolor{gray!15}
\textbf{TAC-CoT} & \textbf{61.2} & \textbf{32.8} & \textbf{67.7} & \textbf{53.9} \\
\bottomrule
\end{tabular}
}

\caption{
Controlled comparisons on the mathematical reasoning benchmarks using Qwen2.5-0.5B-Instruct.
The no-natural-language reasoning control removes natural-language rationales and continuous thought tokens while retaining arithmetic equations.
TokenSkip + Tool provides a concise explicit-CoT control under the tool-enabled setting.
All scores are accuracy (\%).
}
\label{tab:controlled_baselines}
\end{table}

{
Overall, these controlled comparisons indicate that TAC-CoT's improvement is not solely attributable to eliminating natural-language prose or reducing the length of explicit reasoning.
}

\subsection{Additional Full Fine-Tuning Results}\label{sec:finetuning}
We report supplementary results under full fine-tuning for backbone settings not covered in the main results table. Table~\ref{tab:llamafull} shows results for Llama-3.2-1B-Instruct with full fine-tuning, complementing the LoRA-adapted Llama results in Table~\ref{tab:main}. Table~\ref{tab:gpt} reports full fine-tuning results for a GPT-2 backbone. 
Across both settings, \algo consistently improves over the Explicit-CoT agent baseline, further supporting the effectiveness of our method across model families and settings.

\begin{table}[t]
\setlength{\tabcolsep}{6pt}
\renewcommand{\arraystretch}{1.0}
\centering
\resizebox{0.47\textwidth}{!}{

\begin{tabular}{l| ccc | c}
\toprule
 & \multicolumn{4}{c}{\textbf{Math}} \\
 \cmidrule(l{2pt}r{2pt}){2-5}
\multirow{-2.5}{*}{\textbf{Method}}  & {GSM8K} & {GSM-Hard} & {SVAMP} & Average\\
\midrule

Explicit-CoT  & 52.2 & 13.9 & 68.5 & 44.9 \\
\rowcolor{gray!15}
\algo & \textbf{62.0} & \textbf{14.9} & \textbf{68.7} & \textbf{48.5} \\

\bottomrule
\end{tabular}}
\caption{Math accuracy of Explicit-CoT agent and \algo on GSM8K, GSM-Hard, and SVAMP when Llama-3.2-1B-Instruct is fully fine-tuned (no LoRA), reported for reference against the LoRA-adapted Llama numbers in \autoref{tab:main}.}
\label{tab:llamafull}
\end{table}

\begin{table}[t]



\setlength{\tabcolsep}{6pt}
\renewcommand{\arraystretch}{1.0}
\centering
\resizebox{0.47\textwidth}{!}{

\begin{tabular}{l| ccc | c}
\toprule
 & \multicolumn{4}{c}{\textbf{Math}} \\
 \cmidrule(l{2pt}r{2pt}){2-5}
\multirow{-2.5}{*}{\textbf{Method}}  & {GSM8K} & {GSM-Hard} & {SVAMP} & Average\\
\midrule

Explicit-CoT  & 32.8 & 6.6 & 38.2 & 25.9 \\
\rowcolor{gray!15}
\algo & \textbf{47.7} & \textbf{10.8} & \textbf{45.0} & \textbf{34.5} \\

\bottomrule
\end{tabular}}
\caption{Math accuracy of Explicit-CoT agent and \algo on GSM8K, GSM-Hard, and SVAMP under a GPT2 backbone with full fine-tuning, complementing the Qwen and Llama backbones reported in \autoref{tab:main}.}
\label{tab:gpt}
\end{table}

\begin{figure*}[t]
    \centerline{\includegraphics[width=\linewidth]{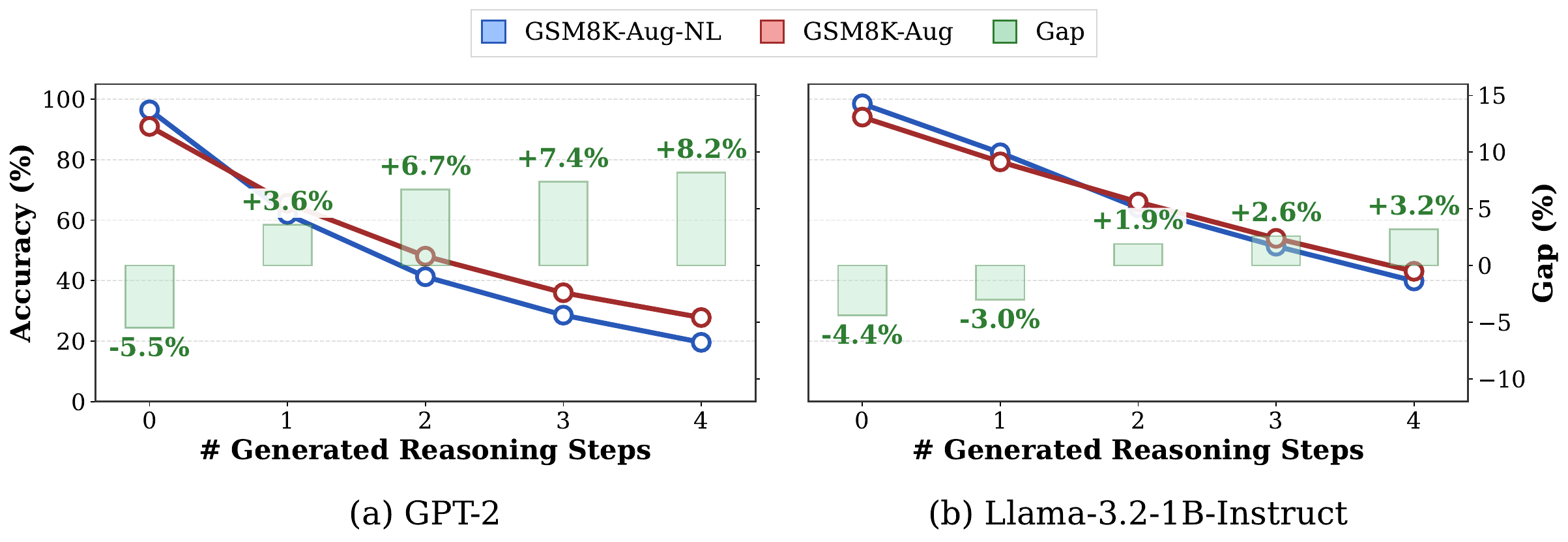}}
    \caption{Effect of reasoning depth on error accumulation.}
    \label{fig:ErrorAcc}
\end{figure*}

\begin{figure}[t]
    \centerline{\includegraphics[width=0.75\columnwidth]{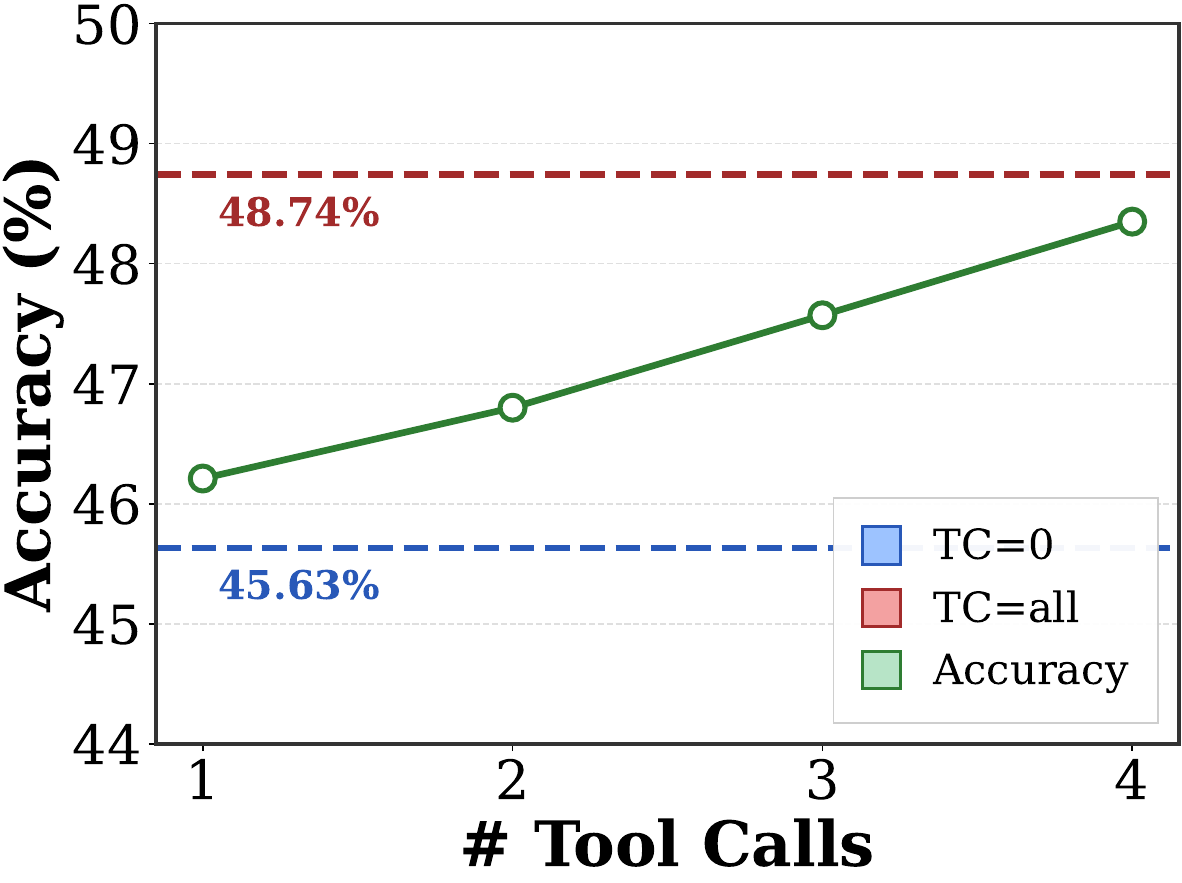}}
    \caption{Impact of the number of tool calls on GSM8K accuracy. TC=0 denotes no tool calling, and TC=all denotes unlimited tool calls.}
    \label{fig:impactoftc} 
\end{figure}

\subsection{Impact of the Tool-Use Budget}\label{sec:tool_use_budget}
To investigate how the number of permitted tool-use affects performance, we conduct an ablation study on GSM8K samples containing four or more equations, since each equation corresponds to a potential calculator invocation. We vary the maximum number of calculator calls from 0 to 4, and also evaluate with unlimited calls. As shown in Figure~\ref{fig:impactoftc}, accuracy improves monotonically as the number of allowed tool calls increases. This suggests that each additional tool call can contribute to mitigate error propagation during multi-step reasoning.

\subsection{Impact of Error Accumulation}\label{sec:error_accumulation}
As shown in Figure~\ref{fig:ErrorAcc}, we trained GPT-2 and Llama-3.2-1B-Instruct on GSM8k-Aug and GSM8K-Aug-NL to analyze how intermediate reasoning errors accumulate as reasoning depth increases. In these datasets, equations are inserted into explicit reasoning. When gold reasoning is provided, GSM8K-Aug-NL-trained models outperform models trained on GSM8K-Aug, which are not trained to generate natural-language rationales. However, this advantage progressively diminishes as the number of generated reasoning steps increases. This trend suggests that, although explicit natural language reasoning can be beneficial under oracle reasoning conditions, error accumulation during intermediate steps can degrade accuracy. The degradation is less pronounced for the larger 1B model, which is consistent with prior observations that larger models are more robust to error accumulation in multi-step reasoning~\cite{recum2026are}.


\begin{table}[t]
\centering
\small
\setlength{\tabcolsep}{6pt}
\resizebox{\linewidth}{!}{
\begin{tabular}{l|l| cc}
\toprule
\textbf{Method} & \textbf{Dataset} & \textbf{\#\,Fwd} & \textbf{ms\,/\,sample} \\
\midrule
\multicolumn{4}{c}{\textcolor{gray!90}{\textbf{\textit{Agentic Math}}}}\\
\midrule
\multirow{4}{*}{\algo}
 & GSM8K     & 43.9 & 545.0 \\
 & GSM-Hard  & 57.0 & 712.2 \\
 & SVAMP     & 26.2 & 335.9 \\
 \noalign{\vskip 2pt}
\cline{2-4}
\noalign{\vskip 2pt}
 & \textbf{Average} & \textbf{42.4} & \textbf{531.0} \\
\cmidrule(lr){1-4}
\multirow{4}{*}{Explicit-CoT}
 & GSM8K     & 109.2 & 1232.2 \\
 & GSM-Hard  & 145.7 & 1673.5 \\
 & SVAMP     & 58.4 & 682.5 \\
 \noalign{\vskip 2pt}
\cline{2-4}
\noalign{\vskip 2pt}
 & \textbf{Average} & \textbf{104.4} & \textbf{1196.1} \\
 \midrule
\multicolumn{4}{c}{\textcolor{gray!90}{\textbf{\textit{Agentic QA}}}}\\
 \midrule
\multirow{4}{*}{\algo}
 & HotpotQA     & 49.4 & 4455.3 \\
 & 2WikiMulti  & 39.8 & 2409.6 \\
 & TriviaQA     & 43.3 & 2697.7 \\
 \noalign{\vskip 2pt}
\cline{2-4}
\noalign{\vskip 2pt}
 & \textbf{Average} & \textbf{44.2} & \textbf{3187.5} \\
\cmidrule(lr){1-4}
\multirow{4}{*}{Explicit-CoT}
 & HotpotQA     & 272.2 & 7765.8 \\
 & 2WikiMulti  & 224.5 & 7372.7 \\
 & TriviaQA     & 235.3 & 7825.0 \\
\noalign{\vskip 2pt}
\cline{2-4}
\noalign{\vskip 2pt}
 & \textbf{Average} & \textbf{244.0} & \textbf{7654.5} \\
\bottomrule
\end{tabular}
}
\caption{
Inference-time comparison between \algo and Explicit-CoT on the Qwen2.5-0.5B-Instruct backbone.
\textbf{ms/sample} denotes the average wall-clock inference latency in milliseconds per sample, measured on a single NVIDIA L40S GPU.
}
\label{tab:latency}
\end{table}
\subsection{Wall-Clock Speedup of \algo}
\label{sec:latency}
We additionally report wall-clock inference time on a single NVIDIA L40S GPU using Qwen2.5-0.5B-Instruct.
As shown in Table~\ref{tab:latency}, \algo consistently reduces both the number of forward passes and end-to-end latency compared with the Explicit-CoT agent.
On average, \algo uses 2.3$\times$ fewer forward passes on the math benchmarks and 5.5$\times$ fewer forward passes on the QA benchmarks.
These reductions also translate into practical wall-clock speedups: \algo runs 2.3$\times$ faster on math benchmarks and 2.4$\times$ faster on QA benchmarks on average.
When isolating model generation time by excluding external overheads, the speedup further increases, reaching up to 3.8$\times$.

These speedups are smaller than the reductions in forward passes because forward-pass count does not translate linearly into end-to-end latency.
In our current implementation, total latency still includes non-optimized execution paths, parallelization overheads, and API-call latency.
Thus, while \algo already provides consistent practical speedups in wall-clock time, the gap between forward-pass reduction and realized latency reduction suggests that further gains are possible with more optimized implementation and execution-level parallelization.

\section{Case Studies}\label{sec:case_study}

\begin{figure*}[t]
\centering\footnotesize

\begin{tcolorbox}[
  title={\textbf{(a) Agentic Math}~(GSM8K)},
  colback=white, colframe=black!50,
  fonttitle=\footnotesize\bfseries,
  boxrule=0.5pt, arc=3pt,
  left=4pt, right=4pt, top=3pt, bottom=3pt
]
\textbf{Q:}~Eliza's rate per hour for the first 40 hours she works each week is \$10.
She also receives an overtime pay of 1.2 times her regular hourly rate.
If Eliza worked for 45 hours this week, how much are her earnings for this week?
\hfill\textbf{Gold:}~\textbf{\$460}

\vspace{3pt}\noindent\rule{\linewidth}{0.3pt}\vspace{3pt}

\begin{cbox}{yellow!25}
\textbf{Gold Trajectory}\\[2pt]
Eliza is entitled to 45\,-\,40\,=\,\texttt{<<45-40=5>>}5 hours overtime pay.
Her hourly rate for the overtime pay is \$10\,x\,1.2\,=\,\$\texttt{<<10*1.2=12>>}12.
So, Eliza will receive \$12\,x\,5\,=\,\$\texttt{<<12*5=60>>}60 for overtime pay.
Her regular weekly earning is \$10\,x\,40\,=\,\$\texttt{<<10*40=400>>}400.
Thus, Eliza will receive a total of \$400\,+\,\$60\,=\,\$\texttt{<<400+60=460>>}\textbf{460} for this week's work.
\end{cbox}

\vspace{2pt}

\begin{cbox}{blue!12}
\textbf{Explicit-CoT (SFT)}\hfill\vwrong~\$940\\[2pt]
Eliza earns \$10\,x\,40\,=\,\texttt{<<10*40=400>>}\$400 for the regular hours she worked.
The overtime pay she receives is \$10\,x\,1.2\,=\,\texttt{<<10*1.2=12>>}\$12 per hour.
So, she earns \colorbox{red!30}{\$12\,x\,\textbf{45}\,=\,\texttt{<<12*45=540>>}\$540} for the overtime.~{\scriptsize$\leftarrow$\textit{should be 5\,h}}
Therefore, Eliza earns \$400\,+\,\$540\,=\,\texttt{<<400+540=940>>}\$940 for this week. The answer is: \textbf{940}
\end{cbox}

\vspace{2pt}

\begin{cbox}{green!15}
\textbf{SIM-CoT}\hfill\vwrong~210\\[2pt]
\LblockN{6}\quad\texttt{[[ANSWER]]}~The answer is: \textbf{210}
\end{cbox}

\vspace{2pt}

\begin{cbox}{red!12}
\textbf{TAC-CoT (Ours)}\hfill\vcorrect~\$460\\[2pt]
\texttt{<<10*40=400>>}\;\LblockN{3}\;
\texttt{<<1.2*10=12>>}\;\LblockN{3}\;
\texttt{<<12*5=60>>}\;\LblockN{3}\;
\texttt{<<400+60=460>>}\;\LblockN{3}\;
\texttt{[[ANSWER]]}~The answer is: \textbf{460}
\end{cbox}

\end{tcolorbox}

\vspace{5pt}

\begin{tcolorbox}[
  title={\textbf{(b) Agentic QA}~(HotpotQA)},
  colback=white, colframe=black!50,
  fonttitle=\footnotesize\bfseries,
  boxrule=0.5pt, arc=3pt,
  left=4pt, right=4pt, top=3pt, bottom=3pt
]
\textbf{Q:}~What island is located 808\,nmi northwest of the capital of Hawaii?
\hfill\textbf{Gold:}~\textbf{Laysan}

\vspace{3pt}\noindent\rule{\linewidth}{0.3pt}\vspace{3pt}

\begin{cbox}{yellow!25}
\textbf{Gold Answer}\quad\textbf{Laysan}\quad{\itshape(No gold trajectory available for HotpotQA.)}
\end{cbox}

\vspace{2pt}

\begin{cbox}{blue!12}
\textbf{Explicit-CoT (SFT)}\hfill\vwrong~O\kern-.1em\textquoteleft ahu\\[2pt]
Thought\,1: I need to find the capital of Hawaii first.\\
Search[\texttt{Hawaii capital}]\;$\to$\;Obs.\,1:~\textit{Could not find. Similar: [\ldots`Honolulu (redirect from Capital of Hawaii)'\ldots]}\\
Thought\,2: The capital of Hawaii is Honolulu. Find an island 808\,nmi northwest of Honolulu.\\
Search[\texttt{Islands near Honolulu}]\;$\to$\;Obs.\,2:~\textit{Could not find. Similar: [\ldots`O\kern-.1em\textquoteleft ahu (redirect from Island of O\kern-.1em\textquoteleft ahu)'\ldots]}\\
Thought\,3: O\kern-.1em\textquoteleft ahu is mentioned. Let me check O\kern-.1em\textquoteleft ahu.\\
Search[\texttt{O\kern-.1em\textquoteleft ahu}]\;$\to$\;Obs.\,3:~\textit{\ldots O\kern-.1em\textquoteleft ahu is 44 miles long from east to west\ldots}\\
Thought\,4--8: Multiple failed searches for the distance between Honolulu and O\kern-.1em\textquoteleft ahu.\\
\colorbox{red!30}{Thought\,9: I recall O\kern-.1em\textquoteleft ahu is approximately 808\,nmi northwest of Honolulu.}\;{\scriptsize$\leftarrow$\textit{factually wrong}}\\
finish[\texttt{O\kern-.1em\textquoteleft ahu}]
\end{cbox}

\vspace{2pt}

\begin{cbox}{green!15}
\textbf{SIM-CoT}\hfill\vwrong~Wake Island\\[2pt]
\textit{(No tool use)}\quad
\LblockN{6}\quad\texttt{[[ANSWER]]}\quad\texttt{<finish>}\colorbox{red!30}{Wake Island}\texttt{</finish>}
\end{cbox}

\vspace{2pt}

\begin{cbox}{red!12}
\textbf{TAC-CoT (Ours)}\hfill\vcorrect~Laysan\\[2pt]
Search[\texttt{Hawaii}]\;$\to$\;Obs.\,1:~\textit{\ldots capital and most populous city\ldots Honolulu\ldots}\;\LblockN{3}\\
Search[\texttt{808 miles northwest of Honolulu}]\;$\to$\\
\quad Obs.\,2:~\textit{Could not find. Similar: [\ldots\underline{Laysan (redirect from Laysan Island)}\ldots]}\;\LblockN{3}\\
Search[\texttt{Laysan}]\;$\to$\;Obs.\,3:~\textit{\ldots Laysan\ldots located \underline{808 nautical miles\ldots northwest of Honolulu}\ldots}\;\LblockN{3}\\
\texttt{[[ANSWER]]}~\textbf{Laysan}
\end{cbox}

\end{tcolorbox}

\caption{%
Case studies on Qwen2.5-0.5B-Instruct.
{\setlength{\fboxsep}{1pt}
\colorbox{yellow!25}{\,Gold\,}
\colorbox{blue!12}{\,Explicit-CoT (SFT)\,}
\colorbox{green!15}{\,SIM-CoT\,}
\colorbox{red!12}{\,TAC-CoT (ours)\,}.
\colorbox{red!30}{\,Red\,}: error step.
}
\textbf{(a)}~Explicit-CoT applies the overtime rate to all 45\,h instead of 5\,h;
SIM-CoT's fixed latent budget yields a wrong answer with no intermediate steps;
TAC-CoT grounds each arithmetic step via calculator calls.
\textbf{(b)}~SIM-CoT hallucinates ``Wake Island'' without retrieval;
Explicit-CoT searches but rationalizes a wrong answer with a false claim (Thought 9);
TAC-CoT spots ``Laysan'' in search suggestions and retrieves the evidence.}

\label{fig:case_study}
\end{figure*}

Figure~\ref{fig:case_study} presents qualitative examples from agentic math and QA using Qwen2.5-0.5B-Instruct.
Explicit-CoT and \algo are both tool-using agents, while SIM-CoT does not expose intermediate tool interactions.

In the math example, Explicit-CoT calls the calculator but applies the overtime rate to all 45 hours instead of the 5 overtime hours.
SIM-CoT also fails without explicit intermediate computation.
\algo grounds each arithmetic sub-step with calculator calls and preserves the correct decomposition of regular pay, overtime hours, and final summation.

In the QA example, SIM-CoT hallucinates ``Wake Island'' without retrieval.
Explicit-CoT searches for evidence but follows an irrelevant O`ahu suggestion and later makes a false factual claim.
\algo instead uses the failed search's similar-results list to identify ``Laysan'' and then retrieves confirming evidence.

\section{Code Details}\label{sec:code_details}

\begin{figure*}[t]
\centering
\small
\begin{minipage}{0.97\textwidth}
\hrule
\vspace{4pt}
\textbf{TAC-CoT Inference} \hfill
(\textit{B}: segment length,\;
 $d$: hidden size,\;
 \textit{kv}: KV-cache)
\vspace{4pt}
\hrule
\vspace{6pt}

\begin{minipage}[t]{0.48\textwidth}
\textbf{(1) Encode prefix $[\texttt{BOS};\, q_{1:n};\, \texttt{[[LATENT]]}]$}
\begin{verbatim}
# full_ids = [BOS] + question_ids
#           + [LATENT_id]
mask = ones_like(full_ids)       # (1, L)
out  = model(input_ids=full_ids,
             attention_mask=mask,
             use_cache=True,
             output_hidden_states=True)
kv = out.past_key_values
h = out.hidden_states[-1][:,-1,:]
c = Proj(h.unsqueeze(1))  # (1,1,d)
\end{verbatim}

\vspace{4pt}
\textbf{(2) Latent iteration loop} \; ($t = 1, \ldots, B$)
\begin{verbatim}
for t in range(B):
  mask = cat([mask, ones(1,1)],
             dim=-1)
  pos = (mask.cumsum(-1)-1)[:,-1:]
  out = model(
    inputs_embeds   = c,  # (1,1,d)
    past_key_values = kv,
    attention_mask  = mask,
    position_ids    = pos,
    use_cache=True,
    output_hidden_states=True)
  h = out.hidden_states[-1][:,-1,:]
  c = Proj(h.unsqueeze(1))
  kv = out.past_key_values
\end{verbatim}

\vspace{4pt}
\textbf{(3) Route: action or answer?}
\begin{verbatim}
mask = cat([mask, ones(1,1)], dim=-1)
pos  = (mask.cumsum(-1)-1)[:,-1:]
out  = model(inputs_embeds=c,
             past_key_values=kv,
             attention_mask=mask,
             position_ids=pos,
             use_cache=True)
kv = out.past_key_values
lg  = out.logits[0, -1]
act = max(lg[action_id], lg[ans_id])
ans = lg[answer_id]

if ans >= act:
  # answer generation (greedy
  #   until </finish> or EOS)
else:
  # -> step (4): action generation
\end{verbatim}
\end{minipage}
\hfill
\begin{minipage}[t]{0.48\textwidth}
\textbf{(4) Action generation and observation injection}
\begin{verbatim}
# greedy generate action tokens
# until </action> stop token

# prepend closing token before obs
inject = [action_close_tok] + obs_ids
mask = cat([mask,
            ones(1,len(inject))], dim=-1)
pos = (mask.cumsum(-1)-1)[:,
                          -len(inject):]
out = model(input_ids=inject,
            past_key_values=kv,
            attention_mask=mask,
            position_ids=pos,
            use_cache=True)
kv = out.past_key_values
# -> back to step (2)
\end{verbatim}

\end{minipage}

\vspace{4pt}
\hrule
\end{minipage}
\caption{%
  Minimal TAC-CoT inference loop.
  $c_t \in \mathbb{R}^d$ is produced by
  $\mathrm{Proj}_\theta$
  (Linear$_{d\to d_p}$ $\to$ GELU $\to$ Linear$_{d_p\to d}$ $\to$ LayerNorm)
  and fed back via \texttt{inputs\_embeds}, bypassing vocabulary projection.
}
\label{fig:inference-pseudocode}
\end{figure*}

Figure~\ref{fig:inference-pseudocode} presents the \algo inference loop code.
We describe the key implementation decisions below.

\paragraph{Continuous thought token injection.}
The continuous thought token $c_t \in \mathbb{R}^d$ is produced by applying a projection head $\mathrm{Proj}_\theta$ to the last hidden state of the transformer: $c_t = \mathrm{Proj}_\theta(h_t)$, where $h_t$ is extracted from the final layer at the last sequence position.
The continuous token is fed back to the model via \texttt{inputs\_embeds}, bypassing the vocabulary embedding lookup entirely.
No vocabulary index is ever assigned to a continuous thought token.

\paragraph{Route selection and observation injection.}
At the end of each latent segment, $c_B$ is fed via \texttt{inputs\_embeds} to obtain route logits.
If the \texttt{[[ANSWER]]} logit exceeds $\ell_{\texttt{[[EXPLICIT]]}}$, the model switches to answer generation;
otherwise, action tokens are generated greedily until a stop token (\texttt{</action>}).
Upon a tool call, the closing action token and the returned observation text are concatenated and processed in a single forward pass, after which the latent iteration loop restarts.


\begin{figure*}[p]
\centering\footnotesize

\begin{tcolorbox}[
  title={\textbf{Prompt for QA Tool-Use Trajectory Construction}},
  colback=white, colframe=black!50,
  fonttitle=\footnotesize\bfseries,
  boxrule=0.5pt, arc=3pt,
  left=3pt, right=3pt, top=2pt, bottom=2pt
]

\begin{promptbox}{black!4}
\textbf{Instruction and Rules}\\[-1pt]
\begin{PromptVerb}
You are a ReAct agent that answers multi-hop questions using Wikipedia.

Process:
1. Think step by step about what information you need.
2. Use exactly ONE action per turn: Search[entity], Lookup[keyword], or Finish[answer].
3. After your action, you will receive an Observation with the result. Then repeat.

Rules:
- Always write "Thought N:" with your reasoning BEFORE "Action N:".
- Use Search[entity] to find Wikipedia articles.
- Use Lookup[keyword] to find specific info within the current article.
- Use Finish[answer] when you have enough information.
- Only one action per turn.
\end{PromptVerb}
\end{promptbox}

\vspace{2pt}

\begin{promptbox}{yellow!15}
\textbf{Few-shot Demonstrations: Examples 1--3}\\[-1pt]
\begin{PromptVerb}
Example 1.
Question: What is the elevation range for the area that the eastern sector of the Colorado orogeny extends into?
Thought 1: I need to search Colorado orogeny, find the area that the eastern sector of the Colorado orogeny extends into, then find the elevation range of the area.
Action 1: Search[Colorado orogeny]
Observation 1: The Colorado orogeny was an episode of mountain building (an orogeny) in Colorado and surrounding areas.
Thought 2: It does not mention the eastern sector. So I need to look up eastern sector.
Action 2: Lookup[eastern sector]
Observation 2: (Result 1 / 1) The eastern sector extends into the High Plains and is called the Central Plains orogeny.
Thought 3: The eastern sector of Colorado orogeny extends into the High Plains. So I need to search High Plains and find its elevation range.
Action 3: Search[High Plains]
Observation 3: High Plains refers to one of two distinct land regions
Thought 4: I need to instead search High Plains (United States).
Action 4: Search[High Plains (United States)]
Observation 4: The High Plains are a subregion of the Great Plains. From east to west, the High Plains rise in elevation from around 1,800 to 7,000 ft (550 to 2,130 m).[3]
Thought 5: High Plains rise in elevation from around 1,800 to 7,000 ft, so the answer is 1,800 to 7,000 ft.
Action 5: Finish[1,800 to 7,000 ft]

Example 2.
Question: Musician and satirist Allie Goertz wrote a song about the "The Simpsons" character Milhouse, who Matt Groening named after who?
Thought 1: The question simplifies to "The Simpsons" character Milhouse is named after who. I only need to search Milhouse and find who it is named after.
Action 1: Search[Milhouse]
Observation 1: Milhouse Mussolini Van Houten is a recurring character in the Fox animated television series The Simpsons voiced by Pamela Hayden and created by Matt Groening.
Thought 2: The paragraph does not tell who Milhouse is named after, maybe I can look up "named after".
Action 2: Lookup[named after]
Observation 2: (Result 1 / 1) Milhouse was named after U.S. president Richard Nixon, whose middle name was Milhous.
Thought 3: Milhouse was named after U.S. president Richard Nixon, so the answer is Richard Nixon.
Action 3: Finish[Richard Nixon]

Example 3.
Question: Which documentary is about Finnish rock groups, Adam Clayton Powell or The Saimaa Gesture?
Thought 1: I need to search Adam Clayton Powell and The Saimaa Gesture, and find which documentary is about Finnish rock groups.
Action 1: Search[Adam Clayton Powell]
Observation 1: Could not find [Adam Clayton Powell]. Similar: ['Adam Clayton Powell III', 'Seventh Avenue (Manhattan)', 'Adam Clayton Powell Jr. State Office Building', 'Isabel Washington Powell', 'Adam Powell', 'Adam Clayton Powell (film)', 'Giancarlo Esposito'].
Thought 2: To find the documentary, I can search Adam Clayton Powell (film).
Action 2: Search[Adam Clayton Powell (film)]
Observation 2: Adam Clayton Powell is a 1989 American documentary film directed by Richard Kilberg. The film is about the rise and fall of influential African-American politician Adam Clayton Powell Jr.[3][4] It was later aired as part of the PBS series The American Experience.
Thought 3: Adam Clayton Powell (film) is a documentary about an African-American politician, not Finnish rock groups. So the documentary about Finnish rock groups must instead be The Saimaa Gesture.
Action 3: Finish[The Saimaa Gesture]
\end{PromptVerb}
\end{promptbox}

\end{tcolorbox}

\caption{Prompt used to construct QA tool-use trajectories. We prompt DeepSeek-V3 as a ReAct-style Wikipedia agent with access to \texttt{Search}, \texttt{Lookup}, and \texttt{Finish}.}
\label{fig:qa_prompt_1}
\end{figure*}

\begin{figure*}[p]
\centering\footnotesize

\begin{tcolorbox}[
  title={\textbf{Prompt for QA Tool-Use Trajectory Construction}~(continued)},
  colback=white, colframe=black!50,
  fonttitle=\footnotesize\bfseries,
  boxrule=0.5pt, arc=3pt,
  left=3pt, right=3pt, top=2pt, bottom=2pt
]

\begin{promptbox}{yellow!15}
\textbf{Few-shot Demonstrations: Examples 4--6}\\[-1pt]
\begin{PromptVerb}
Example 4.
Question: What profession does Nicholas Ray and Elia Kazan have in common?
Thought 1: I need to search Nicholas Ray and Elia Kazan, find their professions, then find the profession they have in common.
Action 1: Search[Nicholas Ray]
Observation 1: Nicholas Ray (born Raymond Nicholas Kienzle Jr., August 7, 1911 - June 16, 1979) was an American film director, screenwriter, and actor best known for the 1955 film Rebel Without a Cause.
Thought 2: Professions of Nicholas Ray are director, screenwriter, and actor. I need to search Elia Kazan next and find his professions.
Action 2: Search[Elia Kazan]
Observation 2: Elia Kazan was an American film and theatre director, producer, screenwriter and actor.
Thought 3: Professions of Elia Kazan are director, producer, screenwriter, and actor. So profession Nicholas Ray and Elia Kazan have in common is director, screenwriter, and actor.
Action 3: Finish[director, screenwriter, actor]

Example 5.
Question: Which magazine was started first Arthur's Magazine or First for Women?
Thought 1: I need to search Arthur's Magazine and First for Women, and find which was started first.
Action 1: Search[Arthur's Magazine]
Observation 1: Arthur's Magazine (1844-1846) was an American literary periodical published in Philadelphia in the 19th century.
Thought 2: Arthur's Magazine was started in 1844. I need to search First for Women next.
Action 2: Search[First for Women]
Observation 2: First for Women is a woman's magazine published by Bauer Media Group in the USA.[1] The magazine was started in 1989.
Thought 3: First for Women was started in 1989. 1844 (Arthur's Magazine) < 1989 (First for Women), so Arthur's Magazine was started first.
Action 3: Finish[Arthur's Magazine]

Example 6.
Question: Were Pavel Urysohn and Leonid Levin known for the same type of work?
Thought 1: I need to search Pavel Urysohn and Leonid Levin, find their types of work, then find if they are the same.
Action 1: Search[Pavel Urysohn]
Observation 1: Pavel Samuilovich Urysohn (February 3, 1898 - August 17, 1924) was a Soviet mathematician who is best known for his contributions in dimension theory.
Thought 2: Pavel Urysohn is a mathematician. I need to search Leonid Levin next and find its type of work.
Action 2: Search[Leonid Levin]
Observation 2: Leonid Anatolievich Levin is a Soviet-American mathematician and computer scientist.
Thought 3: Leonid Levin is a mathematician and computer scientist. So Pavel Urysohn and Leonid Levin have the same type of work.
Action 3: Finish[yes]
\end{PromptVerb}
\end{promptbox}

\vspace{2pt}

\begin{promptbox}{red!8}
\textbf{Input Query}\\[-1pt]
\begin{PromptVerb}
Question: {question}
\end{PromptVerb}
\end{promptbox}

\end{tcolorbox}

\caption{Prompt used to construct QA tool-use trajectories, continued from Figure~\ref{fig:qa_prompt_1}. The few-shot demonstrations follow the ReAct prompt format.}
\label{fig:qa_prompt_2}
\end{figure*}

\end{document}